\documentclass{sn-jnl}

\usepackage[unblind,long,proof]{optional}

\opt{long}{\setlength{\textheight}{232mm}
\usepackage{pslatex}}

\usepackage{amsmath,amssymb,amsthm}
\usepackage[american]{babel} % American English
\usepackage[utf8x]{inputenc}
\usepackage{hyperref,url} % hyperlinks

\usepackage[capitalize,noabbrev]{cleveref}
  \creflabelformat{equation}{#2#1#3}
\theoremstyle{definition}
\newtheorem{defn}{Definition}
\newtheorem{exmp}{Example}
\newtheorem{prop}{Property}
\newtheorem{remk}{Remark}

\newcommand{\seq}[1]{\texttt{#1}}
\newcommand{\nextval}[1]{\underline{#1}}

\usepackage{microtype}
\usepackage{algorithmicx}
\usepackage[numbers]{natbib}
  \opt{blind}{\bibliographystyle{sn-mathphys-num}}
  \opt{long}{\bibliographystyle{unsrtnat}}
\usepackage{booktabs,siunitx}
\usepackage{graphicx,tikz}
\usepackage[figuresright]{rotating}
\usepackage[official]{eurosym}
\usepackage{xcolor}
\usepackage{enumitem}

\newcommand{\head}[1]{\multicolumn{1}{c}{#1}}

\begin{document}

\title{\opt{long}{\bf}Efficient Time-Series Approximation with Linear Recurrent Neural Networks: Architecture Learning and Predictive Power}
\opt{long}{\subtitle{The Power of Linear Recurrent Neural Networks}}

\opt{unblind}{
\author*[1]{\fnm{Frieder} \sur{Stolzenburg}}\email{fstolzenburg@hs-harz.de}
\author[1]{\fnm{Sandra} \sur{Litz}}\email{litz\_sandra@web.de}
\author[2]{\fnm{Olivia} \sur{Michael}}\email{olivia\_michael@hotmail.com}
\author[2]{\fnm{Oliver} \sur{Obst}}\email{o.obst@westernsydney.edu.au}

% Frieder: fstolzenburg@hs-harz.de orcid 0000-0002-4037-2445
% Sandra: litz\_sandra@web.de orcid 0009-0000-8147-4424
% Olivia: olivia\_michael@hotmail.com  orcid 0000-0002-5719-2299
% Oliver: o.obst@westernsydney.edu.au  orcid 0000-0002-8284-2062

\affil[1]{\orgdiv{Faculty of Automation and Computer Sciences}, 
            \orgname{Harz University of Applied Sciences},
            \orgaddress{\street{Friedrichstr.~57-59}, \postcode{38855}~\city{Wernigerode}, \country{Germany}}}

\affil[2]{\orgdiv{Centre for Research in Mathematics and Data Science}, 
            \orgname{Western Sydney University},
             \orgaddress{\street{Locked Bag 1797}, \city{Penrith}~\state{NSW}~\postcode{2751}, \country{Australia}}}
}

\keywords{recurrent neural network, linear activation, time-series analysis, prediction,
dimensionality reduction, approximation theorem, ellipse trajectories}

\abstract{Recurrent neural networks are a powerful means to cope with time series. We show
how autoregressive linear, i.e., linearly activated recurrent neural networks
(LRNNs) can approximate any time-dependent function $f(t)$. The approximation
can effectively be learned by simply solving a linear
equation system; no backpropagation or similar methods are needed. Furthermore,
and this is the main contribution of this paper, the size of an
LRNN can be reduced significantly in one step after inspecting the spectrum
of the network transition matrix, i.e., its eigenvalues, by taking only the most
relevant components. Therefore, in contrast to other approaches, we do not only
learn network weights but also the network architecture.  LRNNs have interesting
properties: They end up in ellipse trajectories in the long run and allow the
prediction of further values and compact representations of functions. We
demonstrate this by several case studies, among them multiple superimposed
oscillators (MSO), robotic soccer (RoboCup), and stock price prediction. LRNNs
outperform the previous state-of-the-art for the MSO task with a minimal number
of units.}

%\begin{keyword}
%\MSC[2010] 15A06 %Linear equations
%\sep 62M10 %Time series, auto-correlation, regression, etc.
%\sep 62M45 %Neural nets and related approaches
%\sep 68T05 %Learning and adaptive systems

%\emph{ACM class:} I.2.6 %Artificial Intelligence - Learning
%\end{keyword}

%\linenumbers
\maketitle

\section{Introduction}
Deep learning generally refers to a class of machine learning algorithms that use a 
cascade of multiple layers of nonlinear processing units for feature extraction 
and transformation \citep{DY14}. The tremendous success of deep learning in 
diverse fields of artificial intelligence, such as computer vision and natural 
language processing, appears to depend on several key components: artificial, 
possibly recurrent neural networks (RNNs) with nonlinearly activated neurons, 
convolutional layers, and iterative training methods such as backpropagation 
\citep{GBC16}. However, which of these components are truly essential for machine 
learning tasks such as time-series analysis?

Research in time-series analysis and hence the modeling dynamics of complex systems 
has a long tradition and remains highly active due to its crucial role in many 
real-world applications \citep{LBE15}, such as weather forecasts, stock quotations, 
comprehension of trajectories of objects and agents, and solving number puzzles 
\citep{RK11,GW13}. Among other benefits, the analysis of time series enables data 
compression, i.e., the compact representation of time series, by a function 
$f(t)$, and prediction of future values.

Numerous studies address these topics using RNNs, particularly variants of 
networks with long short-term memory (LSTM) \citep{HS97}. Recent works in this
area also aim to enhance the performance of RNNs, e.g., HiPPO \citep{GDE+20} and
structured state spaces \citep{GGR22,WKF+25}, which learn representations of signals
over time. There are also more biologically inspired spiking neural networks
\citep{KK+22} that can perfom (simple) delay tasks quite well.

In the following, we consider an alternative, yet simple but very powerful type of RNNs: linear 
recurrent neural networks (LRNNs). We utilize linear activation, which 
allows us to minimize the network size in one step by inspecting the 
eigenvalues of the network transition matrix (cf. \cref{reduce}). 
Unlike other approaches, we do not only learn network weights but also the 
network architecture itself. LRNNs are a computationally efficient complement to more 
demanding approaches, suitable for scenarios where simplicity and rapid deployment are crucial.

The rest of this paper is structured as follows: First, we briefly review related 
works (\cref{related}). We then formally introduce LRNNs as a special and simple 
kind of RNNs together with their properties, including the general network 
dynamics and their long-term behavior (\cref{define}). Afterwards, learning 
LRNNs is explained (\cref{learn}). It is a relatively straightforward procedure 
that allows network size reduction; no backpropagation or gradient-descent 
method is needed. We then discuss results and experiments (\cref{result}), 
before concluding with our findings (\cref{conclude}).

Interestingly, many results for LRNNs presented here can be achieved by 
applying linear algebra and matrix analysis. Our theoretical contributions 
facilitate a highly efficient learning process, previously unexplored in the literature. 
Our approach allows for significant reductions in network size directly after initial training 
(cf. \cref{learn}). In addition, the empirical evaluations, particularly of the MSO 
benchmark and stock price prediction (cf. \cref{mso,stock}) with LRNNs and other 
approaches, such as LSTM networks, demonstrate the practical applicability of our approach. We 
are confident that this work will serve as a valuable resource across various domains, including 
neural engineering, time series prediction, and echo state networks.

\section{Related Works}\label{related}

\subsection{Echo State Networks}

Echo state networks (ESNs) \citep{JH04,Jae07} play a significant role in RNN
research as they provide an architecture and supervised learning principle for
RNNs. They do this by driving a random, large, fixed RNN, called
\emph{reservoir}, with the input signal which then induces in each neuron within
this reservoir network a nonlinear response signal. They combine a desired
output signal with a trainable linear combination of all response signals,
allowing dimensionality reduction by so-called \emph{conceptors} \citep{Jae14,Jae17}.

One variant of ESNs combines several independent (decoupled) smaller networks
\citep{XYH07}. ESN-style (random) initialization has been shown
effective for training RNNs with Hessian-free optimization~\citep{MS11}. The
latter paper addresses the problem of how to effectively train recurrent neural
networks on complex and difficult sequence modeling problems which may contain
long-term data dependencies. This can also be done with LRNNs (cf. MSO
benchmark, \cref{mso}).
Previous work in \citep{Tin18} considers the effect of weight changes in linear symmetric ESNs on
(Fisher) memory of the network. Furthermore, \citet{CW+16} investigate the
asymptotic performance of linear ESNs from a solely theoretical point of view.

\subsection{Recurrent Neural Networks}

Simple RNNs have been proposed originally by \citet{Elm90}. By allowing them to accept sequences
as inputs and outputs rather than individual observations, RNNs extend the
standard feedforward multilayer perceptron networks. As shown in many sequence
modeling tasks, data points such as video frames, audio snippets, and sentence
segments are usually highly related in time. This results in RNNs being used as
the indispensable tools for modeling such temporal dependencies. Linear RNNs and
some of their properties (such as short-term memory) are already investigated in
\citep{WLS94}. Unfortunately, however, it can be a struggle to train RNNs to
capture long-term dependencies \citep{BSF94,PMB13}. This is due to the
gradients vanishing or exploding during backpropagation which in turn makes the
gradient-based optimization difficult.

As a next step,  long short-term memory (LSTM) networks \citep{HS97} became 
a prominent and dominant type of RNN. The expression \emph{long
short-term} refers to the fact that LSTM is a model for the short-term memory
which can last for a long period of time. An LSTM is well-suited to classify,
process and predict time series given time lags of unknown size. They were
developed to deal with the exploding and vanishing gradient problem when
training traditional RNNs. A common LSTM unit is composed of a cell,
an input gate, an output gate, and a forget gate. Each unit type is activated in
a different manner, whereas in this paper we consider completely linearly
activated RNNs.

State-frequency memory (SFM) RNNs \citep{HQ17}  
aim to model the frequency patterns of the temporal sequences.
The key idea of the SFM is to decompose the memory states into different
frequency states. In doing so, they can explicitly learn the dependencies of
both the low and high frequency patterns. As we will see (cf. \cref{mso}),
RNNs in general can easily learn time series that have a constant frequency
spectrum, which may be obtained also by Fourier analysis.

The NoBackTrack algorithm \citep{OTC15} trains RNN parameters in an online, 
memoryless setting
which therefore requires no backpropagation through time. It is also scalable,
thus avoiding the large computational and memory cost of maintaining the full
gradient of the current state with respect to the parameters, but it still uses
an iterative method, namely gradient descent. In contrast to this and other
related works, in this paper we present a method working with linearly activated
RNNs that does not require backpropagation or similar procedures in the learning
phase.
Nevertheless, for feedforward neural networks, the forward-forward algorithm
\citep{Hin22} presents an approach to training models without backpropagation
but, at least in its current form, is only applicable to the classification of
static patterns.

The Legendre memory unit (LMU) \citep{VKE19} is a memory cell for recurrent
neural networks that maintains information across long windows of time using
relatively few resources. It is derived from the linear transfer function for
a continuous-time history of its input signal across a sliding window,
approximated by coupled differential equations, which can implicitly also be
solved by LRNNs (cf. \cref{rem}).
\citet{CSB21} propose an approach to address memorization challenges in RNNs
which puts forward a way between the random encoding in the reservoir paradigm
and the vanishing-gradient prone approach of fully-trained RNNs. The objective
is to train memorization units to maximize their short-term memory capacity,
employing a linear autoencoder for sequences. More recently, deep state-space
models have been shown to perform well on long sequence modeling tasks.
\citet{GGR22} and \citet{OS+23} cover ideas of linear recurrent equations in so-called state
spaces, Jordan-type factorizations, and links to continuous-time differential
equations. Nevertheless, in all these cases, backpropagation is
employed, which is not needed for LRNNs whose network size is reduced
significantly in addition (cf. \cref{reduce}), thus we address the topic of
architecture learning.

Architecture learning, in particular the pruning of neural networks, has been
studied extensively in the literature, with an early survey on pruning algorithms
for neural networks in \citep{Ree93}. More recently, \citet{LAT19} present
a method that prunes irrelevant connections between
neurons for a given task prior to training and is applicable to a variety of
modern neural network models, resulting in sparse networks. Furthermore,
\citet{MM+19} propose a method that estimates the contribution of a neuron to
the final loss and iteratively removes those with smaller scores. In contrast to
this, for LRNNs, the network architecture and hence its size are reduced in one
step by analyzing the network transition matrix (cf. \cref{reduce}).

In more recent works, linear RNNs are considered explicitly: \citet{ZM+25}
establish an explicit connection and a direct architectural match between
Koopman operator approximation and linear RNNs, with the goal of time-series
prediction. \citet{Dur17} considers piecewise linear RNNs and develops a
specific maximum-likelihood estimation scheme for them. Furthermore,
\citet{FCR24} investigate block-diagonal, input-dependent, linear RNNs,
exploring them in the context of regular language modeling. However, none of
these recent works consider network size reduction as done here.

\subsection{Autoregression}\label{regress}

An \emph{autoregressive model} is a representation of a type of random process
\citep{Aka69}. It specifies that the output variable or a vector thereof depends
linearly on its own predecessor values and on a stochastic term (white noise). In
consequence, the model is in the form of a stochastic differential equation as in
general (physical) dynamic systems \citep{CK14}. An LRNN is also linearly
activated, but its output does not only depend on its own predecessor values and
possibly white noise but on the complete state of the possibly big reservoir
whose dynamics is explicitly dealt with. In addition, the size of the network
might be reduced in the subsequent process. We will continue the comparison
between autoregression and LRNNs later (cf. \cref{output}), as we mainly consider
autoregressive tasks in the following.

A popular choice in this context is the autoregressive integrated moving average
(ARIMA) model \citep{HK08,HA13}. A standard ARIMA($p,d,q$) model consists of
autoregression AR($p$) and a moving average MA($q$). The parameter $p$ describes
the history length (lag order) used to predict the time series at time $t$. We
have $f(t) = c_1 f(t-1) + \dots + c_p f(t-p) + e_t$ where $c_1,\dots,c_p$ are
(real-valued) autocorrelation coefficients and $e_t$, the residuals, is
Gaussian white noise. In the moving average process MA($q$), the value $q$
specifies the number of past residuals considered for prediction. An underlying
trend of the time series is modeled by using a drift\opt{long}{, i.e., a constant that extends
the term}. This procedure is particularly well-suited for stationary time
series\opt{long}{, i.e., whose properties do not vary with time}.
Many time series, however, exhibit non-stationary behavior and thus require a
transformation to make them stationary. This is achieved by investigating the
derivatives of the series, with the order of this process given by the parameter
$d$.

State-space time series models, which are a natural bridge between
autoregressive models and LRNNs, have been studied at length \citep{Ham94}.
Autoregressive frameworks are also common in current machine-learning
applications such as large language models (LLMs), e.g., in the generative
pre-trained transformer GPT-3, GPT-4 and related models, by next word prediction
\citep{BM+20,OpenAI24}. The models applied in this context,
however, are very complex (175 billion parameters and much more) and nonlinear.
Related to autoregression is autoencoding of sequences. It has been shown in \citep{PS14}
that linear autoencoders can be used for pre-training of RNNs, while we
establish completely linear RNNs here. Furthermore, \citep{Spe06} gives an
exact closed-form solution for the weights of the linear autoencoder, which is
related to the approximation theorem for LRNNs (cf. \cref{approx}).

\section{Linear Recurrent Neural Networks}\label{define}

RNNs often host several types of neurons, each activated in a different manner
\citep{Elm90,HS97}. In contrast, we understand a homogeneous
interconnected group of standard neurons simply as a (recurrent) neural network here, which may have
arbitrary loops, akin to biological neuronal networks. We adopt a \emph{discrete
time model}, that is, the input and output are represented by a time series and are
processed synchronously and stepwise by the network.

\begin{defn}[time series]
A \emph{time series} is a series of data points in $d$ dimensions
$S(0),\dots,S(n) \in \mathbb{R}^d$ with $d \ge 1$ and $n \ge 0$.
\end{defn}

\begin{defn}[recurrent neural network]
A \emph{recurrent neural network} (RNN) is a directed graph consisting of
altogether $N$ nodes, called \emph{neurons}. $x(t)$ denotes the
\emph{activation} of the neuron $x$ at (discrete) time $t$. We may distinguish
three groups of neurons\opt{long}{ (cf. \cref{net})}:
\begin{itemize}
  \item $N^\mathrm{in}$ \emph{input} neurons (usually without incoming edges)
	whose activation is given by an external source, e.g., a time series,
  \item $N^\mathrm{out}$ \emph{output} neurons (usually without outgoing edges)
	whose activation represents some output function, and
  \item $N^\mathrm{res}$ \emph{reservoir} or \emph{hidden} neurons (arbitrarily
	connected) that are used for auxiliary computations.
\end{itemize}
The sets of input and output neurons are not necessarily disjoint, they may even
be identical (cf. \cref{thedef}). Therefore, in the following, let
$N^\mathrm{in\,out}$ denote the overall number of neurons in the union of both
sets. Obviously it holds $N = N^\mathrm{in\,out}+N^\mathrm{res}$ and
$N^\mathrm{in\,out} \le N^\mathrm{in}+N^\mathrm{out}$.

The edges of the directed graph represent the network connections. They are
annotated with \emph{weights} which are compiled into the \emph{transition
matrix} $W$ of size $N \times N$. An entry~$w_{ij}$ in row~$i$ and column~$j$
denotes the weight of the edge from neuron~$j$ to neuron~$i$. If there is no
connection, then $w_{ij} = 0$. The transition matrix has the form
\begin{equation}\label{matrix}
  W = \left[ \begin{array}{cc}
  \multicolumn{2}{c}{W^\mathrm{out}}\\
  W^\mathrm{in} & W^\mathrm{res}
\end{array} \right]
\end{equation}
containing the following weight matrices:
\begin{itemize}
  \item \emph{input} weights $W^\mathrm{in}$ (weights from the input and
	possibly the output to the reservoir, a matrix of size $N^\mathrm{res} \times N^\mathrm{in\,out}$),
  \item \emph{output} weights $W^\mathrm{out}$ (all weights to the output and
	possibly back to the input, a matrix of size $N^\mathrm{in\,out} \times N$), and
  \item \emph{reservoir} weights $W^\mathrm{res}$ (weights within the reservoir,
	a matrix of size $N^\mathrm{res} \times N^\mathrm{res}$).
\end{itemize}
\end{defn}

\opt{long}{\begin{figure}
  \centering
  \usetikzlibrary {shapes.geometric}
\usetikzlibrary{positioning}
\begin{tikzpicture}[    
    node/.style={draw, circle, minimum size=0.8cm},
    arrow/.style={->, >=latex, thick},
    hidden_layer/.style={draw, ellipse, minimum width=7cm, minimum height=4.5cm}
]

% Input layer
\node[node, label={[label distance=0.5cm,blue]north:Input}] (x1) at (0, 0) {$x_1$};
\node[node] (x2) at (0, -1.5) {$x_2$};
\node[node, label={[label distance=0.5cm]north east:$\mathbf{W}^\text{in}$}] (x3) at (0, -3) {$x_3$};

% Hidden layer
\node[hidden_layer, label={[label distance=0.5cm,red]north:Reservoir}] (hidden) at (5, -1.5) {};
\node[node] (h1) at (3, -1) {$h_1$};
\node[node] (h2) at (3.5, -2.5) {$h_2$};
\node[node] (h3) at (4.5, 0) {$h_3$};
\node[node] (h4) at (5.5, -1.5) {$h_4$};
\node[node] (h5) at (5, -3) {$h_5$};
\node[node, label={[label distance=0.3cm]north west:$\mathbf{W}^\text{res}$}] (h6) at (7, -1) {$h_6$};
\node[node] (h7) at (6.5, -2.7) {$h_7$};

% Output layer
\node[node, label={[label distance=0.5cm,blue]north:Output}] (y1) at (10, -0.75) {$y_1$};
\node[node, label={[label distance=0.1cm]north west:$\mathbf{W}^\text{out}$}] (y2) at (10, -2.25) {$y_2$};

% Edges
\foreach \i in {1,2,3}
    \draw[arrow] (x\i) -- (hidden);

\foreach \j in {1,2}
    \draw[arrow] (hidden) -- (y\j);

% Recurrent connections
\draw[arrow, bend right] (h1) to[out=70,in=135,looseness=0.8] (h3);
\draw[arrow, bend right] (h1) to[out=110,in=135,looseness=0.8] (h5);
\draw[arrow, bend right] (h2) to[out=90,in=90,looseness=0.8] (h4);
\draw[arrow, bend right] (h3) to[out=40,in=160,looseness=0.8] (h7);
\draw[arrow, bend right] (h4) to[out=330,in=195,looseness=0.8] (h7);
\draw[arrow, bend right] (h5) to[out=0,in=200,looseness=0.8] (h6);
\draw[arrow, bend right] (h5) to[out=45,in=130,looseness=0.8] (h2);

\end{tikzpicture}
  \caption{General recurrent neural network. In ESNs, only output weights are
	trained and the hidden layer is also called reservoir.}
  \label{net}
\end{figure}}

Let us now define the \emph{network activity}: The initial
configuration of the neural network is given by a column vector $s$ with $N$
components, called \emph{start vector}. It represents the network state at the
start time $t=t_0$. Because of the discrete time model, the activation
of a (non-input) neuron $x_i$ at time $t+\tau$ (for some time step $\tau>0$) from
the activation at time $t$ of the neurons $x_1,\dots,x_k$ (for some $k \ge 0$),
connected to $x_i$ with the weights $w_{i1},\dots,w_{ik}$, is computed as
follows:
\begin{equation}\label{recur}
	x_i(t+\tau) = g\big(w_{i1}\,x_1(t) +\dots+ w_{ik}\,x_k(t)\big)
\end{equation}
This has to be done simultaneously for all neurons of the network. $g$ is called
\emph{activation function}. Although we will not make use of it,
$g$ may be different for different parts of the network.
Usually, a nonlinear, bounded, strictly increasing, sigmoidal function $g$ is
used, e.g., the logistic function, the hyperbolic tangent ($\tanh$), or the
softplus function \citep[Section~3.10]{GBC16}. In the following, we employ simply
the (linear) identity function, i.e., $g(x)=x$ for all $x$, and can still
approximate arbitrary time-dependent functions (cf. \cref{approx}).

\begin{defn}[linear recurrent neural network]\label{thedef}
A \emph{linear recurrent neural network} (LRNN) is an RNN with the
following properties:
\begin{enumerate}
  \item For the start time, it holds that $t_0=0$ and $\tau$ is constant, usually $\tau=1$.
  \item The initial state $S(0)$ of the given time series constitutes the first
	$d$ components of the start vector $s$.
  \item For all neurons we have \emph{linear activation}, i.e., everywhere $g$
	is the identity.
  \item The weights in $W^\mathrm{in}$ and $W^\mathrm{res}$ are taken
	randomly, independently, and identically distributed from the standard
	normal distribution\opt{long}{, i.e., the Gaussian distribution with mean $\mu = 0$
	and standard deviation $\sigma = 1$,} and remain unchanged all the time,
	whereas the output weights $W^\mathrm{out}$ are learned (cf. \cref{output}).
  \item The spectral radius of the reservoir weights matrix $W^\mathrm{res}$ is
	set to $1$, i.e., $W^\mathrm{res}$ is divided by the maximal absolute
	value of all its eigenvalues. Note that the spectral radius of the
	overall transition matrix $W$ (cf. \cref{matrix}) may still be greater
	than $1$ if required by the application (cf. \cref{continued}).
  \item There is no distinction between input and output but only one (joint)
	group of $N^\mathrm{in\,out} = N^\mathrm{in}=N^\mathrm{out}=d$ input/output neurons. They may
	be arbitrarily connected like the reservoir neurons. We thus can imagine
	the whole network as a big reservoir because the input/output neurons
	are not particularly special.
\end{enumerate}
\end{defn}

LRNNs can run in one of two \emph{modes}: either receiving input or generating
(i.e., predicting) output. In output generating mode, the network runs
autonomously, thus without external input. In this case, \cref{recur} is applied
to all neurons including the input/output neurons. The output from the
previous time step is copied to the input. In input receiving mode, the
activation of every input/output neuron $x$ at time $t$ is always overwritten
with the respective input value at time $t$ given by the time series $S$.

\subsection{Examples}

\begin{exmp}\label{parabola}
The function $f(t) = t^2$ can be realized by an LRNN (in output generating mode)
with three neurons (cf. \cref{examples}(a)). The respective transition matrix
$W$ and start vector $s$ are:
\[ W = \left[ \begin{array}{ccc}
	1 & 2 & 1\\
	0 & 1 & 1\\
	0 & 0 & 1
   \end{array} \right]
   \text{~and~} s = \left[ \begin{array}{c}
	0\\
	0\\
	1
   \end{array} \right]
\]
Consequently, starting at $t=0$ with time step $\tau=1$, we have:
\begin{itemize}
  \item $x_3(0) = 1$, $x_3(t+1) = x_3(t)$, and hence $x_3(t)=1$ in general.
  \item $x_2(0) = 0$, $x_2(t+1) = x_2(t)+x_3(t) = x_2(t)+1$, and hence $x_2(t)=t$.
  \item $x_1(0) = 0$, $x_1(t+1) = x_1(t)+ 2\,x_2(t) + x_3(t) = x_1(t)+ 2\,t +
	1$, and hence $x_1(t) = t^2$ because of the identity $(t+1)^2 = t^2 +
	2\,t + 1$ (cf. first row of the transition matrix $W$).
\end{itemize}
Thus, in the neuron $x_1$, the function $f(t)$ is computed. It is the only
input/output neuron in this case.
\end{exmp}

\begin{exmp}\label{fibonacci}
The Fibonacci series ($0,1,1,2,3,5,8,\dots$) can be defined as follows:
\[ f(t) = \left\{ \begin{array}{ll}
	t, & \text{for~} t=0,1\\
	f(t-1)+f(t-2), & \text{otherwise}
\end{array} \right. \]
It can be realized by an LRNN (in output generating mode) with just two neurons
(cf. \cref{examples}(b)). The respective transition matrix $W$ and start vector
$s$ can be directly derived from the recursive definition of $f$:
\[ W = \left[ \begin{array}{cc}
	0 & 1\\
	1 & 1
   \end{array} \right]
   \text{~and~} s = \left[ \begin{array}{c}
	0\\
	1
   \end{array} \right]
\]
Again, the function $f(t)$ is computed in the only input/output neuron $x_1$. In
the other neuron $x_2$, $f(t+1)$ is generated. There is a closed-form expression
for the Fibonacci series, revealing its exponential growth, known as
\emph{Binet's formula}:
\begin{equation}\label{binet}
   f(t) = \frac{{\lambda_1}^t-{\lambda_2}^t}{\sqrt{5}}
	\text{~with~} \lambda_1 = \displaystyle\frac{1+\sqrt{5}}{2} \approx 1.61803
	\text{~(\emph{golden ratio}) and~} \lambda_2 = 1-\lambda_1
\end{equation}
Interestingly, $\lambda_1$ and $\lambda_2$ are the eigenvalues of the above transition
matrix $W$. Moreover, Binet's formula can be used to create another LRNN to
calculate the Fibonacci series (cf. \cref{examples}(c)). We will come back to
this later (in \cref{continued}).
\end{exmp}

\begin{figure}
  \begin{minipage}[t]{0.32\textwidth}
  \vspace{0pt}
	(a) \begin{tikzpicture}[scale=0.13]
\tikzstyle{every node}+=[inner sep=0pt]
\draw [blue] (5,-25) node {$x_1$};
\draw [red] (20,-21) node {$x_2$};
\draw [red] (20,-37) node {$x_3$};
\draw [black] (4.6,-29.6) circle (3);
\draw (4.6,-29.6) node {$0$};
\draw [black] (4.6,-29.6) circle (2.4);
\draw [black] (24.9,-21.9) circle (3);
\draw (24.9,-21.9) node {$0$};
\draw [black] (24.9,-36.3) circle (3);
\draw (24.9,-36.3) node {$1$};
\draw [black] (24.9,-33.3) -- (24.9,-24.9);
\fill [black] (24.9,-24.9) -- (24.4,-25.7) -- (25.4,-25.7);
\draw (25.4,-29.1) node [right] {$1$};
\draw [black] (27.58,-20.577) arc (144:-144:2.25);
\draw (32.15,-21.9) node [right] {$1$};
\fill [black] (27.58,-23.22) -- (27.93,-24.1) -- (28.52,-23.29);
\draw [black] (27.58,-34.977) arc (144:-144:2.25);
\draw (32.15,-36.3) node [right] {$1$};
\fill [black] (27.58,-37.62) -- (27.93,-38.5) -- (28.52,-37.69);
\draw [black] (22.1,-22.96) -- (7.4,-28.54);
\fill [black] (7.4,-28.54) -- (8.33,-28.72) -- (7.98,-27.78);
\draw (13.68,-25.22) node [above] {$2$};
\draw [black] (22.05,-35.36) -- (7.45,-30.54);
\fill [black] (7.45,-30.54) -- (8.05,-31.27) -- (8.37,-30.32);
\draw (15.76,-32.41) node [above] {$1$};
\draw [black] (5.923,-32.28) arc (54:-234:2.25);
\draw (4.6,-36.85) node [below] {$1$};
\fill [black] (3.28,-32.28) -- (2.4,-32.63) -- (3.21,-33.22);
\end{tikzpicture}
  \end{minipage}
	\hfill
  \begin{minipage}[t]{0.32\textwidth}
  \vspace{0pt}
  	(b) \begin{tikzpicture}[scale=0.13]
\tikzstyle{every node}+=[inner sep=0pt]
\draw (4,-19) node {};
\draw [blue] (5,-25) node {$x_1$};
\draw [red] (26,-25) node {$x_2$};
\draw [black] (5.4,-29.5) circle (3);
\draw (5.4,-29.5) node {$0$};
\draw [black] (5.4,-29.5) circle (2.4);
\draw [black] (26.1,-29.5) circle (3);
\draw (26.1,-29.5) node {$1$};
\draw [black] (23.294,-30.557) arc (-72.74231:-107.25769:25.429);
\fill [black] (23.29,-30.56) -- (22.38,-30.32) -- (22.68,-31.27);
\draw (15.75,-32.2) node [below] {$1$};
\draw [black] (8.189,-28.399) arc (108.02845:71.97155:24.432);
\fill [black] (8.19,-28.4) -- (9.1,-28.63) -- (8.79,-27.68);
\draw (15.75,-26.7) node [above] {$1$};
\draw [black] (28.78,-28.177) arc (144:-144:2.25);
\draw (33.35,-29.5) node [right] {$1$};
\fill [black] (28.78,-30.82) -- (29.13,-31.7) -- (29.72,-30.89);
\end{tikzpicture}
  \end{minipage}
	\hfill
  \begin{minipage}[t]{0.32\textwidth}
  \vspace{0pt}
	(c) \begin{tikzpicture}[scale=0.13]
\tikzstyle{every node}+=[inner sep=0pt]
\draw [blue] (5,-7) node {$x_1$};
\draw [red] (28,-8) node {$x_2$};
\draw [red] (28,-12) node {$x_3$};
\draw [black] (6,-11.4) circle (3);
\draw (6,-11.4) node {$0$};
\draw [black] (6,-11.4) circle (2.4);
\draw [black] (27.9,-3.3) circle (3);
\draw (27.9,-3.3) node {$\lambda_1$};
\draw [black] (27.6,-16.6) circle (3);
\draw (27.6,-16.6) node {$\lambda_2$};
\draw [black] (25.09,-4.34) -- (8.81,-10.36);
\fill [black] (8.81,-10.36) -- (9.74,-10.55) -- (9.39,-9.61);
\draw (18.47,-6.00) node [above] {$\frac{1}{\sqrt{5}}$};
\draw [black] (30.808,-2.614) arc (131.00538:-156.99462:2.25);
\draw (35.23,-5.22) node [right] {$\lambda_1$};
\fill [black] (30.21,-5.19) -- (30.36,-6.12) -- (31.12,-5.47);
\draw [black] (29.982,-14.795) arc (154.88553:-133.11447:2.25);
\draw (34.96,-15.03) node [right] {$\lambda_2$};
\fill [black] (30.48,-17.39) -- (30.99,-18.18) -- (31.42,-17.28);
\draw [black] (24.68,-15.9) -- (8.92,-12.1);
\fill [black] (8.92,-12.1) -- (9.58,-12.78) -- (9.81,-11.8);
\draw (18.47,-13.32) node [above] {$-\frac{1}{\sqrt{5}}$};
\end{tikzpicture}
  \end{minipage}
  \caption{LRNNs for \textbf{(a)} $f(t) = t^2$ and \textbf{(b\,+\,c)} the Fibonacci series
	($0,1,1,2,3,5,8,\dots$) with time step $\tau=1$. In each case, the input/output neuron
	$x_1$ is marked by a double circle. The initial values of the neurons at
	time $t_0=0$ are written in the nodes. The weights are annotated at the
	edges.}
  \label{examples}
\end{figure}

\begin{prop}\label{rem}
LRNNs are well suited to represent differential equations and to solve them
numerically. To see this, consider the homogeneous linear differential equation
\begin{equation}
	\sum_{k=0}^n c_k\,x^{(k)}(t) = 0 \label{diff}
\end{equation}
where $c_k \in \mathbb{R}$ are constant coefficients with $c_n \neq 0$,
$x^{(k)}(t)$ is the $k$-th derivative of the function $x$ with respect to time
$t$, and $n>0$. It can be solved approximately by LRNNs with start vector $s$
satisfying \cref{diff} and the following transition matrix:
\begin{equation} \label{mat}
  W = \left[ \begin{array}{*{5}{c}}
	1 & \tau & 0 & \cdots & 0\\
	0 & 1 & \tau & \ddots & 0\\
	\vdots & \ddots & \ddots & \ddots & 0\\
	0 & \cdots & 0 & 1 & \tau\\
	0 & -\tau \frac{c_0}{c_n} & \cdots & -\tau \frac{c_{n-2}}{c_n} & 1\!-\!\tau \frac{c_{n-1}}{c_n}
      \end{array} \right]
\end{equation}
\end{prop}

\opt{proof}{\begin{proof}
See \cref{proof:rem}.
\end{proof}}

\opt{long}{\begin{exmp}
The exponential function $\exp(t) = e^t$ can be defined by the differential
equation $\dot{x}(t) = x(t)$, i.e., we have $c_0 = 1$ and $c_1 = -1$ in
\cref{diff}. In consequence, according to \cref{rem} and because of $\exp(0)=\dot{\exp}(0)=1$,
the transition matrix $W$ and start vector $s$ of the corresponding LRNN are:
\[ W = \left[ \begin{array}{cc}
	1 & \tau\\
	0 & 1\!+\!\tau
   \end{array} \right]
   \text{~and~} s = \left[ \begin{array}{c}
	1\\
	1
   \end{array} \right]
\]
Induction over time yields immediately $x(t) = \dot{x}(t) = (1+\tau)^{t/\tau}
\approx e^t$ for small $\tau>0$ (according to Euler) as expected.
\end{exmp}}

The strong relationship between RNNs and differential equations is already known
\citep[Section~9]{KB+16} as well as the extraction of eigenvalues to describe
dynamical systems \citep[Section~5]{Str15}. Nevertheless, as we will show in the
rest of this paper, the combination of both provides an effective method for
network size reduction (cf. \cref{reduce}) and therefore seems to be worthwhile
to be considered by the machine learning community in more detail.

\subsection{Network Dynamics}\label{dynamics}

An LRNN runs through network states $f(t)$ for $t
\ge 0$. It holds (in output generating mode)
\[ f(t) = \left\{ \begin{array}{ll}
	s, & t=0\\
	W \cdot f(t-\tau), & \text{otherwise}
\end{array} \right. \]
and hence simply $f(t) = W^t \cdot s$ for $\tau = 1$ (cf. \cref{thedef}).

\begin{prop}\label{jordan}
Let $W = V \cdot J \cdot V^{-1}$ be the Jordan decomposition of the transition
matrix $W$ where $J$ is the direct sum of one or more Jordan blocks, i.e., a
block diagonal matrix formed of Jordan blocks
\[ J_m(\lambda) = \left[ \begin{array}{*{5}{c}}
  \lambda & 1 & 0 & \cdots & 0\\
  0 & \lambda & 1 & \ddots & \vdots\\
  \vdots & \ddots & \ddots & \ddots & 0\\
  \vdots & & \ddots & \lambda & 1\\
  0 & \cdots & \cdots & 0 & \lambda
\end{array} \right] \]
in general with different sizes $m \times m$ and different eigenvalues $\lambda$,
and $V$ is a matrix consisting of the corresponding eigen- and principal column vectors.
Then we have: \[ f(t) = W^t \cdot s = V \cdot J^t \cdot V^{-1} \cdot s \]

If we decompose $V$ into matrices $v$ of size $N \times m$ and the column vector
$V^{-1} \cdot s$ into a stack of column vectors $w$ of size $m$, corresponding
to the Jordan blocks in $J$, then $f(t)$ can be expressed as a sum of vectors $u
= v \cdot J_m(\lambda)^t \cdot w$ where the Jordan block powers are upper
triangular Toeplitz matrices\opt{long}{, i.e., in which each descending diagonal from left
to right is constant,} with:
\begin{equation}\label{power}
  \Big( J_m(\lambda)^t \Big)_{ij} = \binom{t}{j-i}\,\lambda^{t-(j-i)}
	\quad\text{\citep[Section~3.2.5]{HJ13}}
\end{equation}
\end{prop}

\begin{remk}\label{general}
Although the parameter $t$ is discrete, i.e., a nonnegative integer number, the
values of $f(t) = W^t \cdot s$ can also be computed for $t \in \mathbb{R}$ and
are always real. For this, we consider the Jordan block powers from \cref{power}:
\begin{itemize}
  \item The definition of the binomial coefficient $\binom{t}{k} =
	\frac{t\,(t-1)\,\cdots\,(t-k+1)}{k\,(k-1)\,\cdots\,1}$ is applicable for
	real and even complex $t$ and nonnegative integer $k$. For negative $k$,
	we have $\binom{t}{k} = 0$.
  \item For real matrices $W$, there are always complex conjugate eigenvalue
	pairs $\lambda$ and $\overline{\lambda}$ and corresponding complex
	coefficients $c$ and $\overline{c}$ (resulting from the respective matrix $u$ and vector $v$ in
	\cref{jordan}). With $c = |c|\,e^{\mathfrak{i}\psi}$ and
	$\lambda = |\lambda|\,e^{\mathfrak{i}\omega}$, we get $c\,\lambda^t +
	\overline{c}\,\overline{\lambda}{}^t = |c|\,|\lambda|^t \cos(\omega
	t+\psi)$ applying Euler's formula. This obviously is defined for all
	$t \in \mathbb{R}$ and always yields real-valued $f(t)$.
  \item Negative real eigenvalues, i.e., the case $\lambda<0$, should be treated
	in a special way, namely by replacing $\lambda^t$ by $|\lambda|^t \cos(\pi\,t)$.
	Both terms coincide for integer $t$, but only the latter is real-valued
	for all $t \in \mathbb{R}$. The powers of positive real eigenvalues
	$\lambda$ are always positive and real and hence need no special consideration.
\end{itemize}
\end{remk}

A Jordan decomposition exists for every square matrix $W$ \citep[Theorem~3.1.11]{HJ13}.
But if $W$ has $N$ distinct eigenvectors, there is a simpler decomposition,
called \emph{eigendecomposition}. The transition matrix $W$ is
\emph{diagonalizable} in this case, i.e., similar to a diagonal matrix $D$, and
the network dynamics can be directly described by means of the eigenvalues and
eigenvectors of $W$:

\begin{prop}\label{eigen}
Let $W = V \cdot D \cdot V^{-1}$ be the eigendecomposition of the transition matrix $W$ with column
eigenvectors $v_1,\dots,v_N$ in $V$ and corresponding eigenvalues $\lambda_1, \dots, \lambda_N$,
on the diagonal of the diagonal matrix $D$, sorted in decreasing order with
respect to their absolute values. Like every column vector, we can represent the
start vector $s$ as linear combination of the eigenvectors, namely as $s =
x_1 v_1 + \dots + x_N v_N = V \cdot x$ where $x = \big[ x_1 \cdots x_N
\big]^\top$ such that $x = V^{-1} \cdot s$.
Since $W$ is a linear mapping and for each eigenvector $v_k$ with eigenvalue
$\lambda_k$ with $1 \le k \le N$ it holds that $W \cdot v_k = \lambda_k\, v_k$, we
have $W \cdot s = W \cdot (x_1 v_1 + \dots + x_N v_N) = x_1 \lambda_1 v_1 +
\dots + x_N \lambda_N v_N$. Induction over $t$ yields immediately:
\begin{equation}\label{form}
	f(t) = W^t \cdot s = V \cdot D^t \cdot x =
	x_1\,{\lambda_1}^t\,v_1 + \dots + x_N\,{\lambda_N}^t\,v_N
\end{equation}
\end{prop}

\subsection{Real-Valued Transition Matrix Decomposition}\label{real}

For real-valued transition matrices $W$, it is possible to define a
decomposition that, in contrast to the ordinary Jordan decomposition in
\cref{jordan}, solely makes use of real-valued components, adopting the
so-called \emph{real Jordan canonical form} \citep[Section~3.4.1]{HJ13} of the
square matrix $W$\opt{proof}{ (cf. \cref{omi})}. But the
subsequent theorem (\cref{chad}) shows a more general way: The matrix $V$ from
\cref{jordan} is transformed into a real matrix $A$ and, what is more, the start
vector $s$ can be replaced by an arbitrary column vector $y$ with all nonzero
entries.

\begin{prop}\label{chad}
Let $W = V \cdot J \cdot V^{-1}$ be the (real) Jordan decomposition of the
transition matrix $W$ and $s$ the corresponding start vector. Then for all
column vectors $y$ of size $N$ with all nonzero entries, there exists a square
matrix $A$ of size $N \times N$ such that for all $t \ge 0$ we have:
\[ f(t) = W^t \cdot s = A \cdot J^t \cdot y \]
\end{prop}

\opt{proof}{\begin{proof}
See \cref{proof:chad}.
\end{proof}}

\subsection{Long-Term Behavior}\label{ellipse}

Let us now investigate the long-term behavior of an LRNN (run in output generating
mode) by understanding it as an (autonomous) \emph{dynamic system} \citep{CK14,Str15}.
We will see (in \cref{infty}) that the network dynamics may be reduced to a
very small number of dimensions/neurons in the long run. They determine the behavior for $t
\to \infty$. Nevertheless, for smaller $t$, the use of many neurons is important
for computing short-term predictions.

\begin{prop}\label{exponential}
In all of the $N$ dimensions, $f(t) = W^t \cdot s$ grows only polynomially or
single-exponentially in $t$.
\end{prop}

\opt{proof}{\begin{proof}
See \cref{proof:exponential}.
\end{proof}}

In fact, LRNNs can model polynomials (cf. \cref{parabola}, parabola), general
single-exponential functions such as the Fibonacci series (cf. \cref{fibonacci}),
multiple superimposed oscillators (cf. \cref{exmp}), and many more (cf.
\cref{approx}). For this, the overall transition matrix $W$ may have (a)~a
spectral radius greater than $1$ and (b)~many eigenvalues (more than two) with
absolute value $1$. Nevertheless, it is interesting to investigate a special
case, namely a \emph{pure random reservoir} where both conditions do not hold:

\begin{prop}\label{infty}
Consider an LRNN solely consisting of a random reservoir whose transition matrix
$W^\mathrm{res}$ (a)~is completely real-valued, (b)~has an eigendecomposition
$W^\mathrm{res} = V \cdot D \cdot V^{-1}$ (as in~\cref{eigen})
with unit spectral radius, and thus (c)~all eigenvalues are distinct (which is
almost always, i.e., with probability close to $1$, true for random matrices),
together with a completely real-valued random start vector $s$ with unit norm.
Then, almost all terms $x_k\,{\lambda_k}^t\,v_k$ in \cref{form} vanish for large
$t$ because for all eigenvalues $\lambda_k$ with $|\lambda_k| < 1$ we have
$\lim\limits_{t \to \infty} {\lambda_k}^t = 0$. Although a general real matrix
can have more than two complex eigenvalues which are on the unit circle, for a
pure random reservoir as considered here, almost always only the (largest)
eigenvalues $\lambda_1$ and possibly $\lambda_2$ have the (maximal) absolute
values $1$. In consequence, we have one of the following cases:
\begin{enumerate}
  \item $\lambda_1 = +1$. In this case, the network activity contracts to one
	point, i.e., to a \emph{singularity}: $\lim\limits_{t \to \infty}
	f(t) = x_1\,v_1$
  \item $\lambda_1 = -1$. For large $t$ it holds that $f(t) \approx
	x_1\,(-1)^t\,v_1$. This means we have an \emph{oscillation} in this
	case. The dynamic system alternates between the two points $\pm x_1\,v_1$.
  \item $\lambda_1$ and $\lambda_2$ are two (properly) complex eigenvalues with
	absolute value $1$. Since $W^\mathrm{res}$ is a real-valued matrix, the two
	eigenvalues as well as the corresponding eigenvectors $v_1$ and $v_2$
	are complex conjugate with respect to each other. Thus, for large $t$, we
	have an \emph{ellipse} trajectory
	\[
		f(t) \approx x_1\,{\lambda_1}^t\,v_1 + x_2\,{\lambda_2}^t\,v_2
		= \tilde{V} \cdot \tilde{D}^t \cdot \tilde{x}
	\]
	where $\tilde{V} = \big[ v_1\;v_2 \big]$, $\tilde{D} = \left[
	\begin{array}{cc} \lambda_1 & 0 \\ 0 & \lambda_2 \end{array} \right]$,
	and $\tilde{x} = \left[ \begin{array}{c} x_1 \\ x_2 \end{array}
	\right]$.
\end{enumerate}
\end{prop}

\begin{figure}[b]
  \centering
    (a) \raisebox{-35mm}{\includegraphics[scale=0.21]{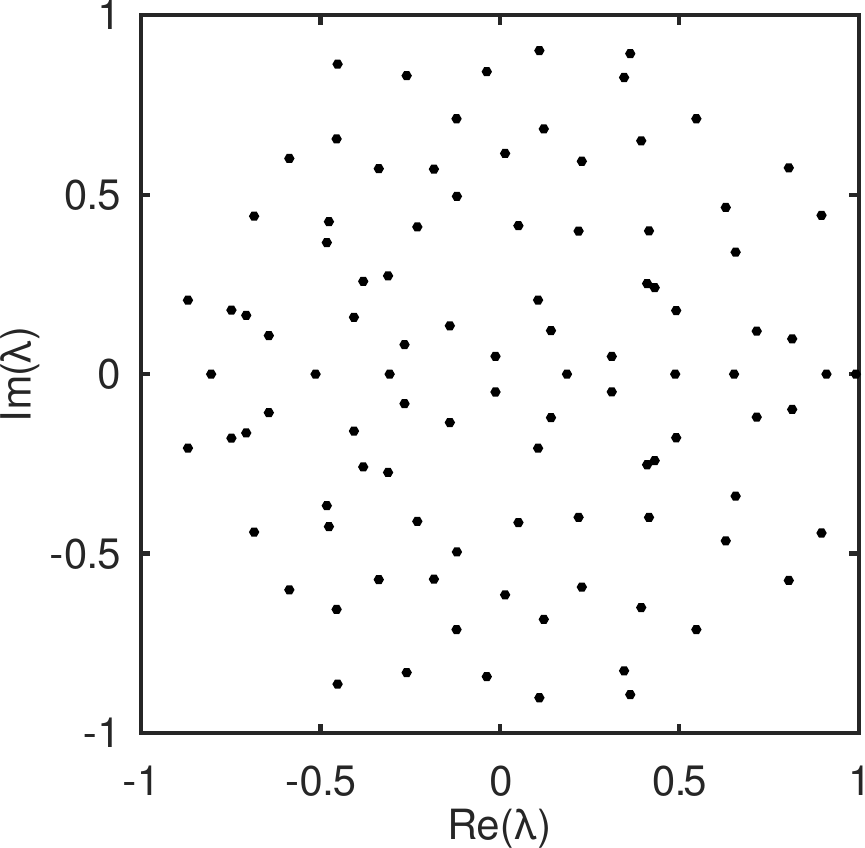}} \hfill
    (b) \raisebox{-35mm}{\includegraphics[scale=0.21]{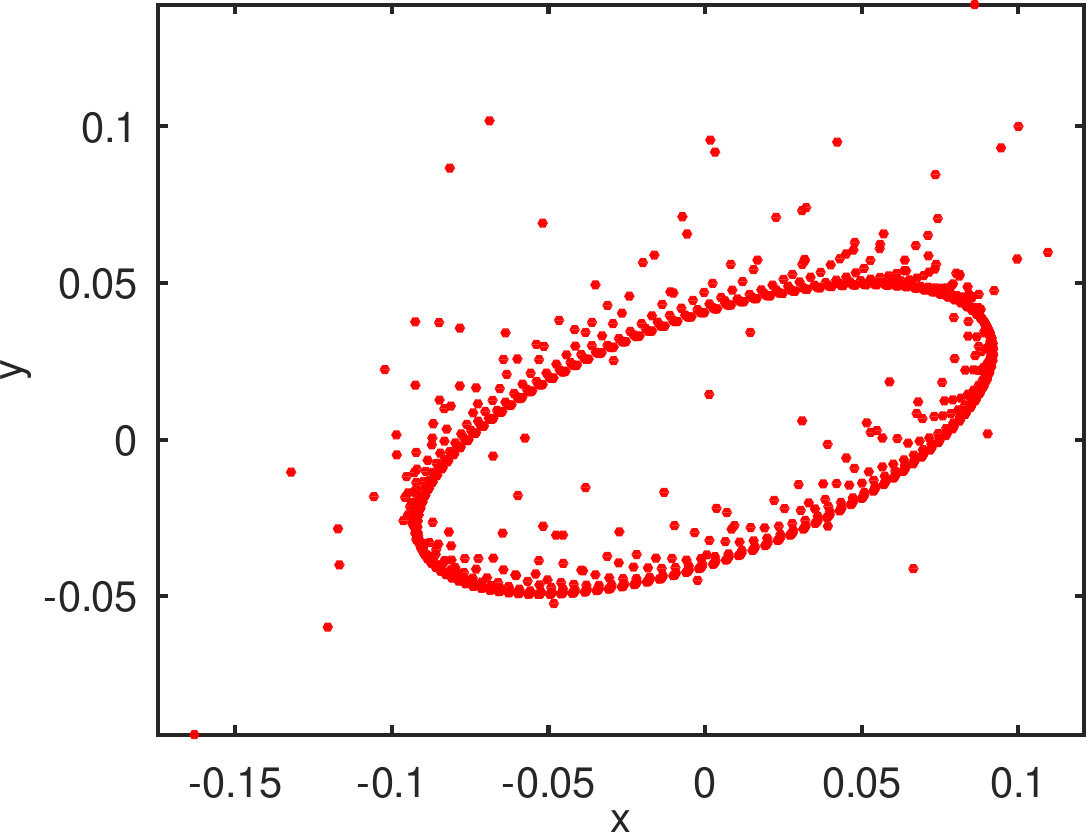}} \hfill
    (c) \raisebox{-35mm}{\includegraphics[scale=0.21]{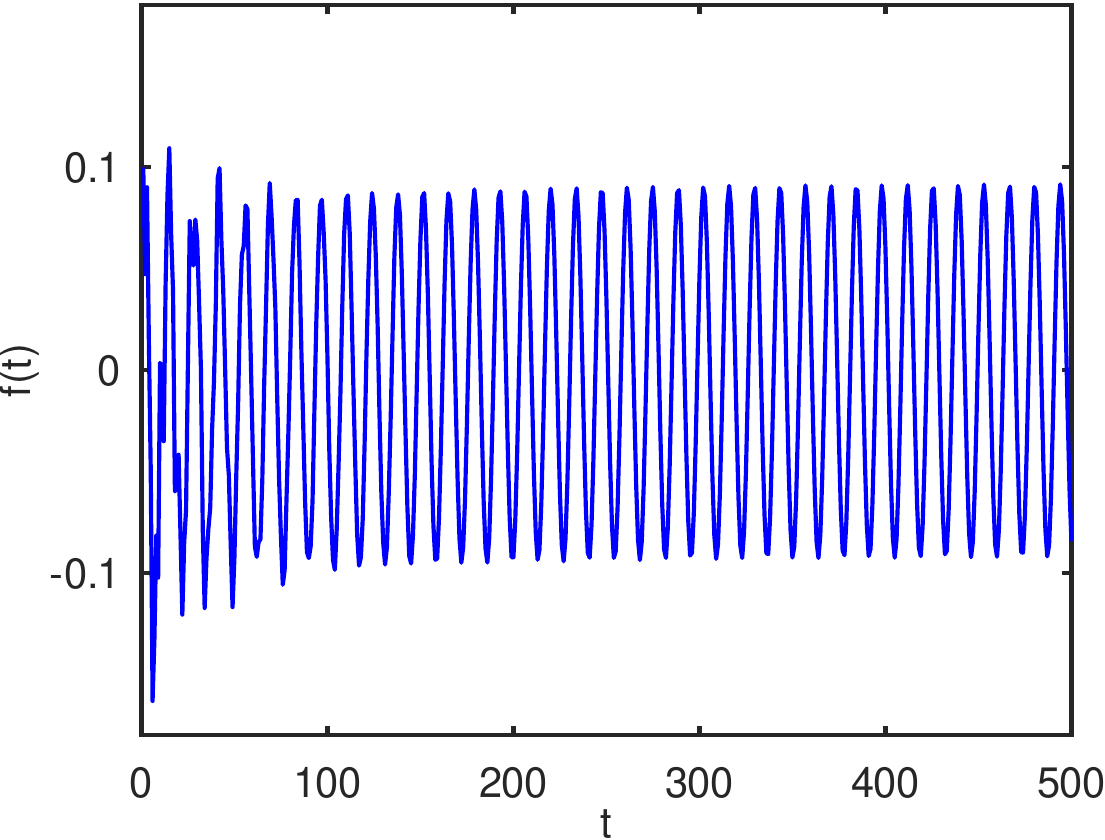}}
  \caption{Dynamic system behavior of a pure random reservoir with unit spectral
	radius, with $N^\mathrm{res} = 100$ neurons:
    \textbf{(a)}~Eigenvalue spectrum of the reservoir matrix $W^\mathrm{res}$ with
	complex conjugate eigenvalue pairs in the complex plane.
    \textbf{(b)}~Visualization of $f(t)$ by planar projection. In the long run, we get an
	ellipse trajectory, thus only two dimensions (cf. \cref{twodim}).
    \textbf{(c)}~Projected to one (arbitrary) dimension, we have pure sinusoids with one
	single angular frequency for large $t$, sampled in large steps.}
  \label{ell}
\end{figure}

\begin{figure}
 \centering
 \includegraphics[width=0.6\textwidth]{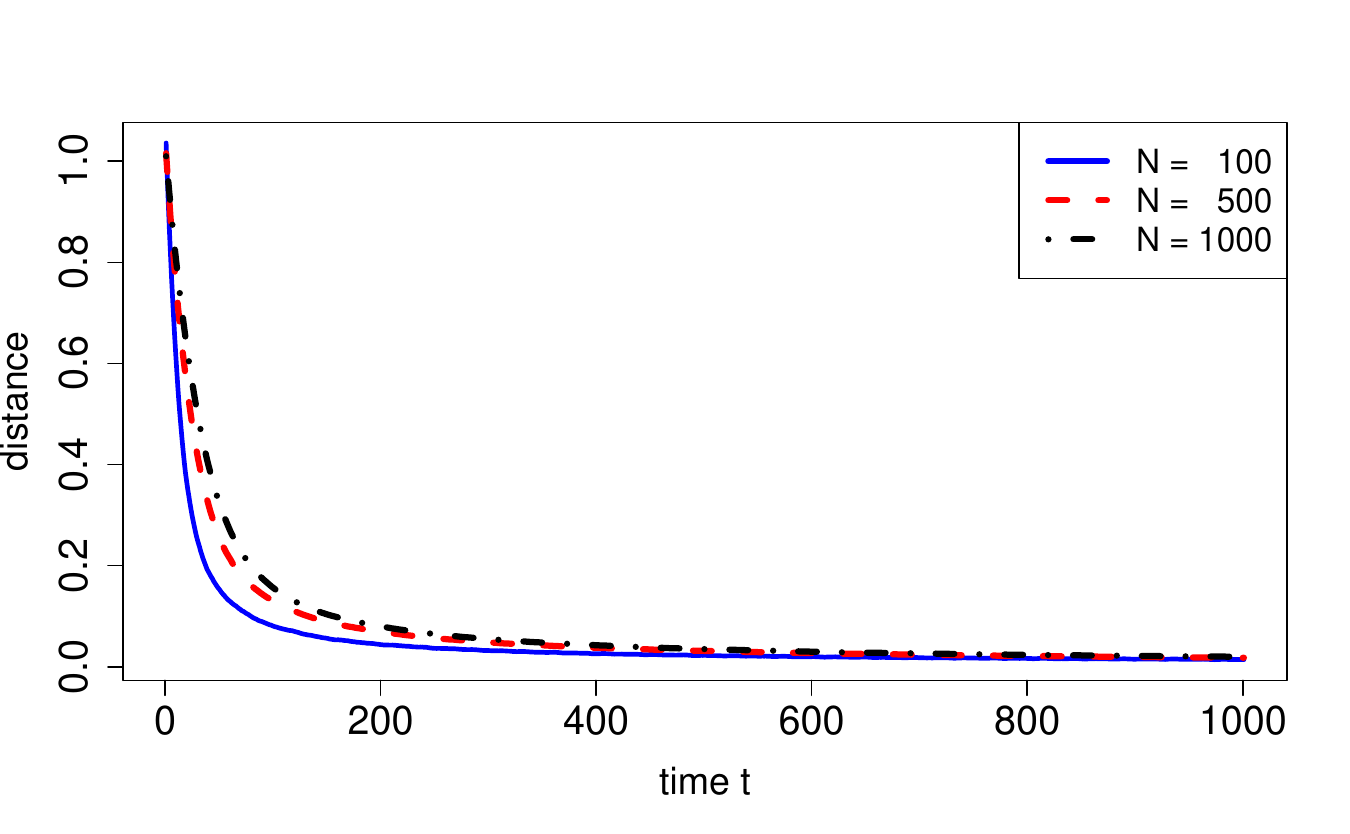} % asymptot1 for solid lines, 0 for dashed
  \caption{Asymptotic behavior of pure random reservoirs with unit spectral radius:
	The (Euclidean) distance between the actual value $f(t)$
	(according to \cref{form}) and its approximation by the final ellipse
	trajectory (\cref{twodim}) is almost zero already after a few hundred
	steps. The figure shows the distances for $N^\mathrm{res}=100$ (solid/blue), $N^\mathrm{res}=500$
	(dashed/red), and $N^\mathrm{res}=1000$ (dotted/black) random reservoir neurons,
	starting with a random vector of unit length, averaged over 1000 trials.}
  \label{asymptot}
\end{figure}

We can build a matrix $\hat{D}$, similar to $\tilde{D}$ but completely
real-valued (cf.~\cref{real}) which states the ellipse rotation. Furthermore,
the rotation speed can be derived from the eigenvalue $\lambda_1$ as follows: In each step
of length $\tau$, there is a rotation by the angle $\omega\tau$ where $\omega$
is the angular frequency, which can be determined from the equation $\lambda_1 =
|\lambda_1|\,e^{\mathfrak{i}\omega\tau}$ (cf. \cref{general}). The
two-dimensional ellipse trajectory can be stated by two (co)sinusoids then:
$f(t) = \big[ a\,\cos(\omega\,t) ~~ b\,\sin(\omega\,t) \big]^\top$ where $a,b >
0$ are half the width and height of the ellipse. Applying the addition theorems
of trigonometry, we get:
\begin{eqnarray*}
	f(t+\tau) & = &
	\left[ \begin{array}{c} a\,\cos\!\big(\omega\,(t+\tau)\big) \\ b\,\sin\!\big(\omega\,(t+\tau)\big) \end{array} \right]  = 
	\left[ \begin{array}{c}
		a\,\big(\!\cos(\omega\,t) \cos(\omega\,\tau) - \sin(\omega\,t) \sin(\omega\,\tau) \big) \\
		b\,\big(\!\sin(\omega\,t) \cos(\omega\,\tau) + \cos(\omega\,t) \sin(\omega\,\tau) \big)
	\end{array} \right] \\ & = &
	\underbrace{\left[ \begin{array}{cc}
		\cos(\omega\,\tau) & -a/b \sin(\omega\,\tau) \\
		b/a \sin(\omega\,\tau) & \cos(\omega\,\tau)
	\end{array} \right]}_{\displaystyle\hat{D}}\,\cdot\,f(t)
\end{eqnarray*}
From this, we can read off the desired ellipse rotation matrix $\hat{D}$ as
indicated above. Due to \cref{chad}, there exists a (two-dimensional) start
vector $y$ and a transformation matrix $A$ such that
\begin{equation}\label{twodim}
	f(t) \approx A \cdot \hat{D}^t \cdot y
\end{equation}
for large $t$. Every LRNN with many neurons can thus be approximated by a simple
network with at most two neurons. The output values lie on an ellipse in
general, thus in only two dimensions. Nonetheless, in the beginning, i.e., for
small $t$, the dynamics of the system is not that regular (cf. \cref{ell}). But
although \cref{infty} states only the asymptotic behavior of random LRNNs with
unit spectral radius, interestingly the network dynamics converges relatively
fast to the final ellipse trajectory: The (Euclidean) distance between the
actual value $f(t)$ (according to \cref{form}) and its approximation by the
final ellipse trajectory (\cref{twodim}) is almost zero already after a few
hundred steps (cf. \cref{asymptot}). Of course this depends on the eigenvalue
distribution of the transition matrix \citep{TV10}. So the long-term behavior
may be different for transition matrices other than pure random reservoirs.

The long-term behavior of LRNNs is related to that of ESNs. For the latter,
usually the activation function is $\tanh$ and the spectral radius is smaller
than $1$. Then reservoirs collapse because of $|\!\tanh(z)| \le
|z|$ for all $z \in \mathbb{R}$, but the convergence may be rather slow.
Nonetheless it guarantees contractivity and hence for any fixed input (not just
the origin) the system converges to a unique fixed point. This leads
to the so-called \emph{echo state property} \citep{MJ13}: Any random initial
state of a reservoir is forgotten such that, after a washout period, the current
network state is a function of the driving input. In contrast to ESNs, LRNNs
have linear activation and a spectral radius of exactly~$1$ (cf. \cref{thedef}).
But as we have just shown, there is a similar effect in the long run: The
network activity reduces to at most two dimensions -- independent of the initial
state of the network.

\section{Learning LRNNs}\label{learn}

Functions can be learned and approximated by LRNNs in two steps: First, as for
ESNs \citep{JH04}, we only learn the output weights $W^\mathrm{out}$ (cf.
\cref{output}). The input weights
$W^\mathrm{in}$ and reservoir weights $W^\mathrm{res}$ are arbitrary random
values and remain unchanged (cf. \cref{thedef}). Nevertheless, in order to
obtain better numerical stability during the computation, they are adjusted as
follows:
\begin{itemize}
  \item Because of the linear activation, the spectral radius of the reservoir
	weights matrix $W^\mathrm{res}$ is set to $1$ (cf. \cref{thedef}).
	Otherwise, with increasing $t$, the values of $f(t) = W^t \cdot s$
	explode if the spectral radius is greater or vanish if the spectral
	radius is smaller than $1$ (cf. \cref{ellipse}). In consequence, the
	overall learning procedure behaves rather stable.
  \item Furthermore, we initialize the reservoir neurons such that the reservoir
	start vector $r$ (with $N^\mathrm{res}$ components) has unit norm by setting:
	\[ r = \frac{1}{\sqrt{N^\mathrm{res}}} \cdot \big[ 1 \cdots 1 \big]^\top \]
	It is part of the (overall) start vector $s = \left[ \begin{array}{c}
	S(0) \\ r \end{array} \right]$ (cf. \cref{define}).
  \item We employ fully connected graphs\opt{long}{, i.e., all, especially the
	reservoir neurons are connected with each other,} because the
	connectivity has nearly no influence on the best reachable performance
	\citep{KLB12}.
\end{itemize}
Second, if possible, we reduce the network size (cf. \cref{reduce}). This often
leads to better generalization and avoids overfitting. Thus, in contrast to many
other approaches, the network architecture is changed during the learning
process, and we do not do this by incremental derivation from the original
network but in only one step.

\subsection{Learning the Output Weights}\label{output}

To learn the output weights $W^\mathrm{out}$, we run the input values from the
time series $S(0),\dots,S(n)$ through the network (in input receiving mode),
particularly through the reservoir. This means, we build the sequence of
corresponding reservoir states $R(0),\dots,R(n)$ where the reservoir start
vector $r$ (see above) in principle can be chosen arbitrarily but with all nonzero
entries (cf. \cref{chad}):
\begin{equation}\label{reserv}
	R(t_0) = r \text{~and~} R(t+\tau) \opt{long}{=
	\left[ W^\mathrm{in} ~~ W^\mathrm{res} \right] \cdot
	\left[ \begin{array}{c} S(t) \\ R(t) \end{array} \right]}
	= W^\mathrm{in} \cdot S(t) + W^\mathrm{res} \cdot R(t)
\end{equation}
We want to predict the next input value $S(t+\tau)$, given the current input and
reservoir states $S(t)$ and $R(t)$. To achieve this, we comprise
all but the last input and reservoir states in one matrix $X$ with:
\begin{equation}\label{Xin}
    X = \left[ \begin{array}{ccc}
	S(0) & \cdots & S(n-1)\\
	R(0) & \cdots & R(n-1)
	\end{array} \right]
\end{equation}
Each output value shall correspond to the respective next input value $S(t+\tau)$.
Therefore, we compose another matrix
\begin{equation}\label{Yout}
	Y^\mathrm{out} = \big[ S(1)\ \cdots\ S(n) \big]
\end{equation}
consisting of the next values of the time series $S$ to be predicted
where the first value $S(0)$ clearly has to be omitted because it has no
predecessor value. We compute $Y^\mathrm{out}(t) = S(t+\tau)$ from $X(t)$ by
assuming a linear dependency:
\begin{equation}\label{linear}
	Y^\mathrm{out} = W^\mathrm{out} \cdot X
\end{equation}
Its solution can easily be determined as $W^\mathrm{out} = Y^\mathrm{out}/X$,
where $/$ denotes right matrix division, i.e., the operation of solving a linear
equation system, possibly applying the least squares method in case of an
overdetermined system, as implemented in many scientific programming languages
such as Matlab \citep{HH17} or Octave \citep{EB+17}. Prediction of further values
is now possible (in output generating mode) as follows:
\begin{equation}\label{predict}
	\left[ \begin{array}{c} S(t+\tau) \\ R(t+\tau) \end{array} \right]
	= W \cdot \left[ \begin{array}{c} S(t) \\ R(t) \end{array} \right]
	\text{~with~} W \text{~as in \cref{matrix}}
\end{equation}

\begin{prop}[treatment of multiple sequences]\label{multi}
It is also possible to learn from multiple sequences at once. For this, let
several time series $S_1,\dots,S_K$ in $d$ dimensions with (not necessarily
identical) lengths $n_1,\dots,n_K$ be given. For each $S_k$ with $1 \le k \le
K$, we determine:
\begin{itemize}
  \item the sequence of corresponding reservoir states $R_k$ (according to
	\cref{reserv}), taking always the same reservoir start vector $r$,
  \item the corresponding input matrix $X_k$ (according to \cref{Xin}), and
  \item the corresponding predicted output matrix $Y^\mathrm{out}_k$ (according to \cref{Yout}).
\end{itemize}
We aggregate the input and output matrices to $X = \big[ X_1 \cdots X_K \big]$
and $Y^\mathrm{out} = \big[ Y^\mathrm{out}_1 \cdots Y^\mathrm{out}_K \big]$ with
$n_1+\,\cdots\,+n_K$ columns each. Solving the linear matrix equation
$Y^\mathrm{out} = W^\mathrm{out} \cdot X$ (identical with \cref{linear})
finally yields the output weight matrix $W^\mathrm{out}$.
\end{prop}

This first phase of the learning procedure is related to a linear
\emph{autoregressive model} \citep{Aka69}. However, one important
difference to an autoregressive model is that for LRNNs the output does not only
depend on its own previous values and possibly white noise but on the complete
state of the possibly big reservoir whose dynamics is explicitly dealt with in
the reservoir matrix $W^\mathrm{res}$. The reservoir effectively allows us to do
arbitrary auxiliary computation such that any (nonlinear) function $f(t)$ can
be approximated by an LRNN (cf.~\cref{approx}).

\subsection{An Approximation Theorem}

\begin{prop}\label{approx}
From a function $f(t)$ in $d \ge 1$ dimensions, let a
series of function values $f(t_0),\dots,f(t_n)$ be given. Then there is an
LRNN with the following properties:
\begin{enumerate}
  \item It runs exactly through all given $n+1$ function values, i.e., it approximates $f(t)$.
  \item It can effectively be learned by the LRNN learning procedure (\cref{output}) employing
	\begin{equation}\label{ineq}
	  N^\mathrm{res} \ge n-N^\mathrm{in\,out}
	\end{equation}
	reservoir neurons.
\end{enumerate}
\end{prop}

\opt{proof}{\begin{proof}
See \cref{proof:approx}.
\end{proof}}

Therefore, at least in theory, any time-dependent function $f(t)$ can be
interpolated, i.e., exactly approximated on the given function values and
continued on input other than nonnegative integer numbers (cf. \cref{general}),
although clearly not every function can be implemented by LRNNs, in particular
functions increasing faster than single-exponential (cf. \cref{exponential})
such as $2^{2^t}$ (double-exponential) or $t\/!$ (factorial function). Also in
practice, the LRNN learning procedure performs rather well (cf. \cref{result}).

Nevertheless, the matrix $X$ may be ill-conditioned for long input sequences, because the reservoir state
sequence as part of the matrix $X$ reduces to at most two dimensions for large
$t$, independent of the number of reservoir neurons (cf. \cref{ellipse}). Hence,
the rank of the matrix $X$ may not be maximal and consequently \cref{linear} may not
be solvable numerically in practice (although we may have an equation system with
the same number of equations and unknowns). A simple increase of the number of
reservoir neurons does not help much.

However, one could learn not only the output weights $W^\mathrm{out}$ as in ESNs
but the complete transition matrix $W$: For this, we employ a random reservoir
state sequence matrix $\big[R(0) \cdots R(n) \big]$ with $N^\mathrm{res}$
reservoir neurons, considered as additional input. If all elements of this
matrix are random numbers, independently and identically distributed from the
standard normal distribution, its rank is almost always maximal, because the
original and the additional (random) inputs are throughout linearly independent
of each other. This procedure therefore avoids the ill-conditioning of the
matrix $X$. We then just have to solve the linear matrix equation $Y = W \cdot
X$ (cf. \cref{linear}) with
\[ Y = \left[ \begin{array}{ccc}
	S(1) & \cdots & S(n)\\
	R(1) & \cdots & R(n)
\end{array} \right] \]
and $X$ as in \cref{Xin}. By this, the input and reservoir weights
$W^\mathrm{in}$ and $W^\mathrm{res}$ are learned, not only the output weights
$W^\mathrm{out}$. But our experiments indicate that this procedure is less
reliable than the one with given, i.e., predefined random input and reservoir
weights and unit spectral radius for the reservoir (cf. \cref{output}), at least
if the whole (and not only a part of) the reservoir is learned from additional
(random) input.

The topic of learning all weights in the matrix $W$ is investigated
in \citep{PDW13} for ESNs with nonlinear activation function in the reservoir.
However, for LRNNs, the given input and reservoir weights $W^\mathrm{in}$ and
$W^\mathrm{res}$ together with the learned output weights $W^\mathrm{out}$
already provide the best approximation of the function $f(t)$. There is no need
to learn the input and reservoir weights, simply because LRNNs are completely
linearly activated RNNs (including the reservoir). If one tries to learn
$W^\mathrm{in}$ and $W^\mathrm{res}$ taking not only the output time series $S$
but additionally the reservoir state time series $R$ into account, then exactly
the given input and reservoir weights are learned if \cref{ineq} holds. Only
with nonlinear activation there is a learning effect.

Although LRNNs rely on linear activation functions, they can represent
nonlinear functions $f(t)$. In fact, LRNNs are well suited to represent
differential equations (cf. \cref{rem}), which lead to highly nonlinear
functions. Any exponential, trigonometric and polynomial functions can be
modeled (cf. \cref{jordan}). There is no restriction to cyclic or periodic time
series. Any time-dependent function $f(t)$ can be exactly approximated (cf.
\cref{approx}), including functions with regime shifts. This is also true for
chaotic functions such as random number series or the Mackey-Glass time series
\citep{MG77,JH04}. However, long-term prediction is not that accurate for these
types of time series.

\begin{remk}[scope and generalization to nonlinear activation]
This paper focuses on time-dependent functions $f(t)$, where the output evolves
autonomously over time. Such functions are central to modeling dynamic systems
and are naturally handled by LRNNs with purely linear activation. In this
setting, nonlinearities are not required: The network dynamics, governed by the
transition matrix, already permit the approximation of a broad class of
functions, including polynomials, exponentials, and oscillatory signals (cf.
\cref{approx}). 

A different problem is to compute general functions $f(s)$, i.e., mappings from
arbitrary input vectors $s$ to output values. In that context, nonlinear
activation functions may offer benefits. If the activation function $g$ is
monotonic and invertible, the LRNN learning procedure introduced so far (\cref{output}) can, in
principle, be extended. For this, $g$ has to be applied component-wise to
\cref{reserv} while running the input through the reservoir and to
\cref{linear}. The solution of the resulting equation $Y^\mathrm{out} =
g\big(W^\mathrm{out} \cdot X\big)$ is then just $W^\mathrm{out} =
g^{-1}\big(Y^\mathrm{out}\big)/X$.

However, the network size reduction procedure introduced next (\cref{reduce})
relies on linear activation and is therefore not applicable when nonlinear
activation is used. In such cases, approximation may still be possible, but the
reduction method would no longer preserve precision because (the proof of)
\cref{chad} requires the commutativity of matrix multiplication with Jordan
matrices $J$, which does not hold any longer if nonlinear activation is
employed.
\end{remk}

\cref{approx} is related to the \emph{universal approximation theorem} for
feedforward neural networks \citep{Hor91}. It states that a (non-recurrent)
network with a linear output layer and at least one hidden layer activated by a
nonlinear, sigmoidal function can approximate any continuous function on a
closed and bounded subset of the $\mathbb{R}^n$ from one finite-dimensional
space to another with any desired nonzero amount of error, provided that the
network is given enough hidden neurons \citep[Section~6.4.1]{GBC16}. Since RNNs
are more general than feedforward networks, the universal approximation theorem
also holds for them \citep{MNM02}. Any measurable function can be
approximated with a (general) recurrent network arbitrarily well in probability
\citep{Ham00}.

\opt{long}{\begin{remk}
Any time series $S(0),\dots,S(n)$ can be generated by employing a backward shift
matrix, i.e., a binary matrix with $1$s on the subdiagonal and $0$s elsewhere
\citep[Section~0.9.7]{HJ13}, as transition matrix $W$ and $s = \big[ S(0) \cdots
S(n) \big]^\top$ as start vector. But such a network clearly would have no
ability to generalize to future data. Fortunately, this does not hold for a
transition matrix $W$ learned by the procedure in \cref{output}. Furthermore,
the eigenvalue spectrum of the backward shift matrix is empty, whereas that of
the learned $W$ is not, which is important for network size reduction
introduced in \cref{reduce}.
\end{remk}}

\subsection{Network Size Reduction}\label{reduce}

To approximate a function exactly for sure, we need a large number
$N^\mathrm{res}$ of reservoir neurons (cf. \cref{approx,ineq}). It is certainly a
good idea to lower this number. One could do this by simply taking a smaller
number of reservoir neurons, but then a good approximation cannot be guaranteed.
In what follows, we therefore reduce the dimensionality of the transition matrix $W$ in a
more controlled way -- after learning the output weights. Our procedure of
dimensionality reduction leads to smaller networks with sparse connectivity. In
contrast to other approaches, we do not learn the new network architecture by
incremental derivation from the original network, e.g., by removing unimportant
neurons or weights, but in one step by inspecting the eigenvalues of the
transition matrix.

For ESNs, dimensionality reduction is considered, too, namely by means of
so-called \emph{conceptors} \citep{Jae14,Jae17,KOS21b}. These are special matrices which
restrict the reservoir dynamics to a linear subspace that is characteristic for
a specific pattern. However, as in principal component analysis (PCA) \citep{Jol11}, conceptors
reduce only the spatial dimensionality of the point cloud of the given data. In
contrast to this, for LRNNs, we reduce the transition matrix $W$ and hence take
also into account the temporal order of the data points in the time series. By
applying insights from linear algebra, the actual network size can be reduced
and not only the subspace of computation as with conceptors.

\begin{prop}\label{relevan}
By \cref{chad}, the function $f(t) = W^t \cdot s$ can be rewritten by
means of the Jordan matrix of the transition matrix $W$ as $A \cdot J^t
\cdot y$, where the start vector can be chosen as nonzero constant, e.g., $y =
\big[ 1 \cdots 1 \big]^\top$. Furthermore, by \cref{jordan}, $f(t)$ can be
expressed as a sum of vectors $u = v \cdot J_m(\lambda)^t \cdot w$ where $w$ is
constant because it is part of the start vector~$y$. Then it follows from
\cref{infty} that for large $t$ the contribution of a Jordan component
vanishes if $\|v\| \approx 0$ and/or $|\lambda| \ll 1$.

In consequence, we omit all Jordan components causing only small errors, until a
given threshold is exceeded. The error $E$ of a network can be estimated by the
root-mean-square error (RMSE) normalized to the number of all sample components
between input $x$ and predicted output $y$:
  \[ \mathrm{RMSE}(x,y) = \sqrt{\frac{1}{n} \sum_{t=1}^n \big\|x(t)-y(t)\big\|^2} \]
We shall omit all network components corresponding to Jordan blocks
$J_m(\lambda)$ with smallest errors as long as the
RMSE is below a given threshold $\theta$. Network components that are not
omitted are considered \emph{relevant}. Thus, from $A$, $J$, and
$y$ (according to \cref{chad}), we successively derive reduced matrices
$A$ and $J$ and the vector $y$ as follows:
\begin{itemize}
  \item Reduce $A$ to the rows corresponding to the input/output components and
	the columns corresponding to the relevant network components.
  \item Reduce $J$ to the rows and columns corresponding to the relevant
	network components.
  \item Reduce $y$ to the rows corresponding to the relevant network components.
\end{itemize}
\end{prop}

Note that the dimensionality reduction does not only lead to a smaller number of
reservoir neurons but also to a rather simple network structure: The transition
matrix $J$ (which comprises the reservoir weights $W^\mathrm{res}$ of
the reduced network) is a sparse matrix with nonzero elements only on the main
and immediately adjacent diagonals. Thus, the number of connections is in
$O(N)$, i.e., linear in the number of reservoir neurons, not quadratic -- as in
general.

\cref{proc} summarizes the overall learning procedure for LRNNs including
network size reduction. It has been implemented by the authors in Octave. Note
that, although the Jordan matrix $J$ (cf. \cref{jordan}) may
contain eigenvalues with multiplicity greater than $1$, Octave does not always
calculate exactly identical eigenvalues then. Therefore, we cluster the
computed eigenvalues as follows: If the distance in the complex plane between
eigenvalues is below some given small threshold $\delta$, they are put into the
same cluster which eventually is identified with its centroid. Thus it is a kind
of single linkage clustering \citep{GR69}. In our approach, the Jordan
components are considered in decreasing order of their relevance, sorted by the
RMSE that their omission causes. Therefore, they cannot be removed by
mistake, provided that a reasonable threshold $\theta$ is taken. However, in our
network size reduction procedure (which is not a simple pruning), we use a
binary search (cf. \cref{proc}). It handles a large group of similarly relevant
eigenvalues in one step. This speeds up the procedure considerably, although it
is a kind of approximation.
The complete implementation of the learning procedure together with some case
studies (cf. \cref{result})\opt{unblind}{ is publicly available at
\url{http://github.com/OliverObst/decorating/}.}\opt{blind}{ will be made
available upon acceptance.}

\begin{figure}
\newcommand{\commt}[1]{\%~#1\\}%
\begin{algorithmic}[1]\State
	\commt{$d$-dimensional function $f$, given sampled, as time series $S$, and start vector $s$}
	$S = \big[ f(0) \cdots f(n) \big]$\\
	$s = \left[ \begin{array}{c} S(0) \\ r \end{array} \right]$ where
		$r = \frac{1}{\sqrt{N^\mathrm{res}}} \cdot
		\big[ \underbrace{1 \cdots 1}_{N^\mathrm{res} \text{~times}} \big]^\top$\newline
	\\
	\commt{random initialization of input and reservoir weights}
	$W^\mathrm{in} = \mathrm{randn}(N^\mathrm{res},d)$\\
	$W^\mathrm{res} = \mathrm{randn}(N^\mathrm{res},N^\mathrm{res})$
		normalized to unit spectral radius\newline
	\\
	\commt{learn output weights by linear regression}
	$X = \left[W^t \cdot s\right]_{t=0,\dots,n}$ \text{~\% run in input receiving mode}\\
	$Y^\mathrm{out} = \big[ S(1)\ \cdots\ S(n) \big]$\\
	$W^\mathrm{out} = Y^\mathrm{out}/X$\newline
	\\
	\commt{transition matrix and its decomposition}
	$W = \left[ \begin{array}{cc}
		\multicolumn{2}{c}{W^\mathrm{out}}\\
		W^\mathrm{in} & W^\mathrm{res}
	\end{array} \right]$\\
	$J = \mathrm{jordan\_matrix}(W)$ with components $k$ sorted in decreasing order\\
	\quad with respect to $\mathrm{Error}(J_{\langle 1,\dots,k-1,k+1,\dots,K \rangle})$\\
	\quad where $K = \#$\,Jordan components in $J$\newline
	\\
	\commt{network size reduction (with binary search)}
	$L = 1$ \text{~\% left index border}\\
	$R = K$ \text{~\% right index border}\\
	while ($L \neq R$)\\
	\quad $M = \left\lfloor\frac{L+R}{2}\right\rfloor$\\
	\quad if $\mathrm{Error}(J_{\langle 1,\dots,M \rangle}) < \theta$\\
	\qquad then $R = M$\\
	\qquad else $L = M+1$\\
	return\big($M$\big)\newline
	\\
	\commt{subroutine $\mathrm{Error}(J_I)$}
	\commt{\quad compute error for Jordan matrix reduced to indexed components}
	\quad reduce $J$ to components indexed by $I$\\
	\quad $y = \big[ 1 \cdots 1 \big]^\top$\\
	\quad $Y = \left[J^t \cdot y\right]_{t=0,\dots,n}$ \text{~\% run in output generating mode}\\
	\quad $A = X/Y$ with rows restricted to input/output dimensions\\
	return\big($\mathrm{RMSE}(S,A \cdot Y)$\big)
\end{algorithmic}
\caption{Pseudocode for learning LRNNs including network size reduction. A
binary search algorithm is employed for determining the relevant network
components with smallest errors. For this, the network components are sorted by
their RMSE. The program returns the number $M$ of relevant components in the
Jordan matrix $J$ (line 24). The subroutine $\mathrm{Error}(J_I)$ (lines 25-31)
computes the error of the predicted output for the Jordan matrix reduced to the
components indexed by $I$.}
\label{proc}
\end{figure}

\begin{exmp}\label{continued}
Let us illustrate the LRNN learning procedure (\cref{proc}) with the Fibonacci
series (\cref{fibonacci}). We start with the first values of the Fibonacci
series $f(0),\dots,f(n)$ as input $S$ (lines 1-3) and generate a random
reservoir of size $N^\mathrm{res}$ (lines 4-6). After learning the output
weights $W^\mathrm{out}$ (lines 7-10) and decomposing the resulting transition
matrix $W$ (lines 11-15), the network size reduction procedure (lines 16-24)
often yields minimal networks with only two reservoir neurons representing
\[
    f(t) \overset{\text{\cref{chad}}}{=} A \cdot J^t \cdot y \text{~with~}
	A \approx \left[ \frac{1}{\sqrt{5}} ~~ -\!\frac{1}{\sqrt{5}} \right] \text{~and~}
	J \approx \left[ \begin{array}{cc}
		\lambda_1 & 0\\
		0 & \lambda_2
   	\end{array} \right] \text{~for~}
	y = \left[ \begin{array}{c}
	1\\
	1
   \end{array} \right]
\]
where the eigenvalues $\lambda_1$ and $\lambda_2$ are as in Binet's formula
(\cref{binet}). For instance, for $N^\mathrm{res}=n=30$ and precision threshold
$\theta=0.001$, we obtain minimal networks in 32\% of the cases from $100$
trials. They belong to the best networks with respect to their RSME. Thus,
by employing a standard validation procedure, the LRNN in \cref{examples}(c)
actually can be derived numerically by the LRNN learning procedure with network
size reduction.
\end{exmp}

Note that the spectral radius of the reservoir weights matrix $W^\mathrm{res}$
remains $1$ all the time (cf. \cref{thedef}). However, after learning the output
weights $W^\mathrm{out}$, the spectral radius of the overall transition matrix
$W$ (according to \cref{matrix}) and hence of the matrix $J$ may
be greater than $1$ if the function $f(t)$ to be modeled is exponentially
increasing. This obviously holds for the Fibonacci series (because of
$\lambda_1>1$).

\subsection{Complexity and Generalization of the Procedure}

\begin{remk}\label{complexity}
In both learning steps, it is possible to employ any of the many available fast
and constructive algorithms for linear regression and eigendecomposition
\citep{DDH07}. Therefore, the time complexity is just $O(N^3)$ for both output
weights learning and network size reduction. In theory, if we assume that
the basic numerical operations such as $+$~and~$\cdot$ can be done in constant
time, the asymptotic complexity is even a bit better. In practice, however, the
complexity depends on the bit length of numbers in floating point arithmetic,
of course, and may be worse hence. The size of the learned network is in $O(N)$
(cf. \cref{reduce}).
\end{remk}

Note that, in contrast, feedforward networks with three threshold neurons
already are NP-hard to train \citep{BR92}. This results from the fact
that the universal approximation theorem for feedforward networks differs from
\cref{approx} because the former holds for multivariate functions and
not only time-dependent input. In this light, the computational complexity of
$O(N^3)$ for LRNNs does not look overly expensive. It dominates the
overall time complexity of the whole learning procedure because it is not
embedded in a time-consuming iterative learning procedure (such as backpropagation)
as in other state-of-the-art methods.

\opt{long}{\begin{remk}
We observe that most of the results presented in this paper still
hold if the transition matrix $W$ contains complex numbers. This means in
particular that also complex functions can be learned (from complex-valued time
series) and represented by LRNNs (\cref{approx}).
Nonetheless, the long-term behavior of networks with a random complex transition
matrix $W$ differs from the one described in \cref{ellipse} because then
there are no longer pairs of complex conjugate eigenvalues.
\end{remk}}

\section{Experiments}\label{result}

In this section, we demonstrate evaluation results for learning and predicting
time series, approximating them by a function $f(t)$ represented by an LRNN,
for several tasks. We consider the following benchmarks: multiple superimposed
oscillators (MSO), number puzzles, robot soccer simulation, and predicting stock
prices. All experiments are performed with a program written by the authors in
Octave \citep{EB+17} (cf. \cref{reduce}). \opt{long}{There is also a reimplementation
in Python available (see \url{http://github.com/OliverObst/lrnn}).} Let us start
with an example that illustrates the overall method.

\begin{figure}
	\centering	
        \hfill %
	\includegraphics[width=0.45\textwidth]{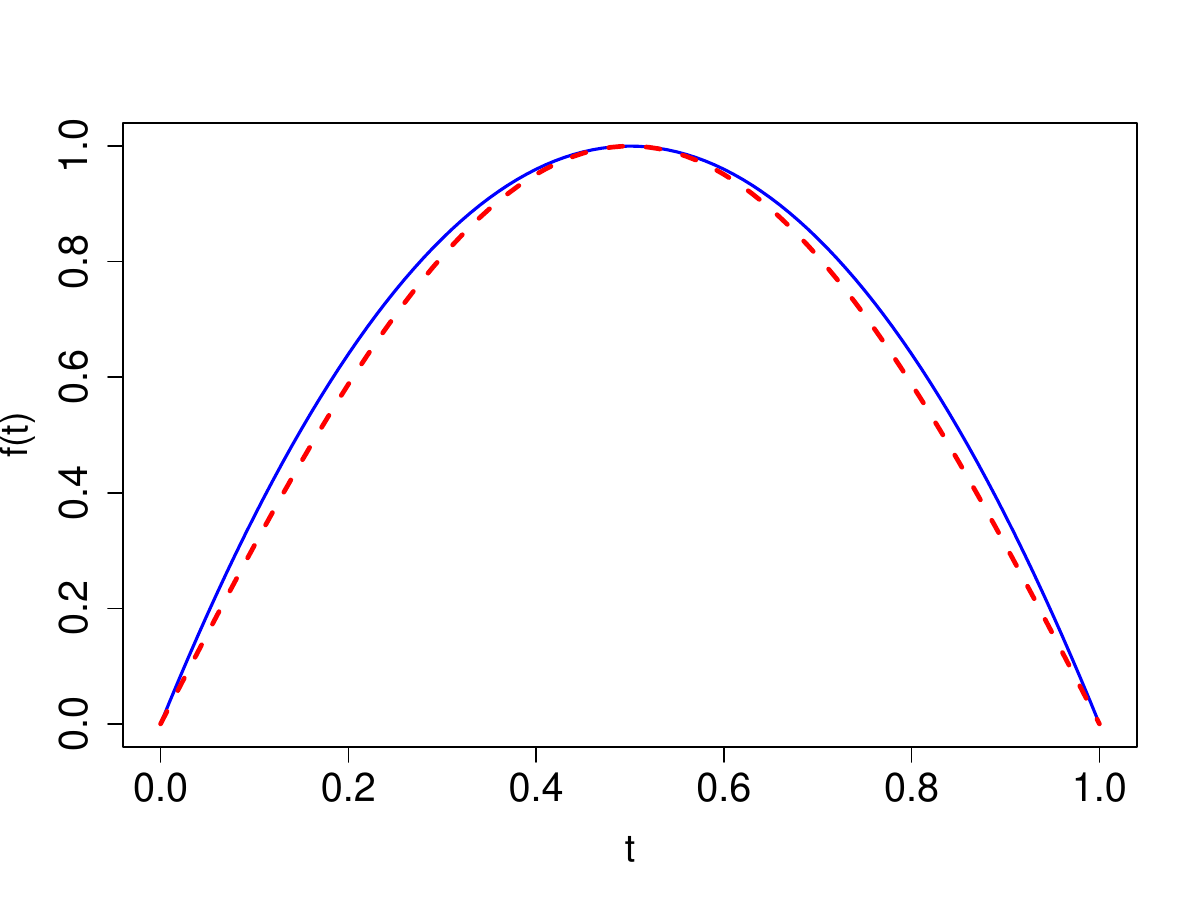}
        \hfill %
	\includegraphics[width=0.45\textwidth]{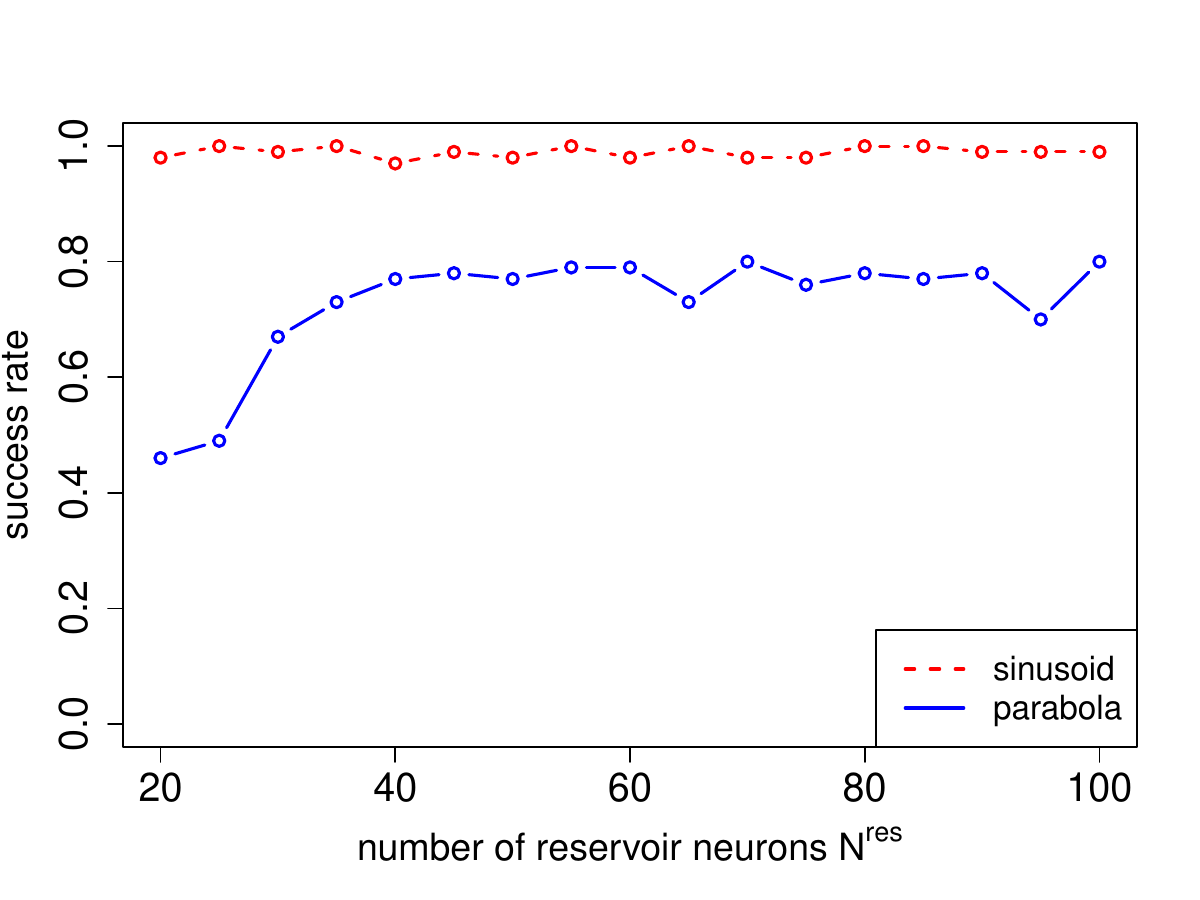}
	\hfill
	\caption{\textbf{Left}: Graphs for \cref{quest}: a parabola and a sinusoid.
	The question is: which one is which? Both can be learned and
	distinguished by LRNNs from the visually similar positive parts of the
	respective graphs, i.e., function values for $t \in [0, 1]$. In this
	interval, all values of the parabola (solid/blue) are greater than or equal
	to those of the sinusoid (dashed/red).
	\textbf{Right}: For \cref{quest}, how often LRNNs of minimal size are
	learned after network size reduction, i.e., with $N_1 = 3$ neurons for
	the parabola and $N_2 = 2$ neurons for the sinusoid? The diagram shows
	the success rate of the learning procedure in this regard as a function
	of the number of reservoir neurons $N^\mathrm{res}$ before network size
	reduction (for 100 trials). Networks of minimal size are learned
	starting already with $N^\mathrm{res} = 40$ reservoir neurons in about
	77\% (parabola, solid/blue) or 99\% (sinusoid, dashed/red) of the
	trials.}
    \label{parasin}
\end{figure}

\begin{exmp}\label{quest}
The graphs of the functions $f_1(t) = 4\,t\,(1-t)$ (parabola) and $f_2(t) =
\sin(\pi\,t)$ (sinusoid) look rather similar for $t \in [0,1]$ (cf.
\cref{parasin}~Left). Can both functions be learned and distinguished from each other
by our LRNN learning procedure (cf. \cref{learn})?
\end{exmp}

To investigate this, we sample both graphs for $t \in [0,1]$ with time step
$\tau=0.01$. After that, we learn the output weights $W^\mathrm{out}$ (cf.
\cref{output}), starting with a large enough reservoir consisting of up to
$N^\mathrm{res} = 100$ neurons (cf. \cref{approx}). Finally, we reduce the
size of the overall transition matrix $W$ with precision threshold $\theta =
0.01$ and cluster threshold $\delta = 0.03$ (cf. \cref{reduce}). Minimal LRNNs
consist of $N_1 = 3$ neurons for
the parabola (cf. \cref{parabola}) and $N_2 = 2$ neurons for the sinusoid
(cf. \cref{ellipse}). The networks of minimal size are learned already with
$N^\mathrm{res} = 40$ reservoir neurons before network size reduction in about
77\% (parabola) or 99\% (sinusoid) of the trials (cf. \cref{parasin}~Right).
Learning the parabola is more difficult because the corresponding transition
matrix $W$ (cf. \cref{parabola}) has no proper eigendecomposition according
to \cref{eigen} but only a Jordan decomposition according to \cref{jordan}.

\subsection{Multiple Superimposed Oscillators}\label{mso}

\begin{exmp}\label{exmp}
Multiple superimposed oscillators (MSO) count as difficult benchmark problems
for RNNs \citep{KLB12,SW+07}. The corresponding time
series is generated by summing up several (pure) sinusoids. Formally
it is described by
	\[ S(t) = \sum\limits_{k=1}^K \sin(\alpha_k\,t) \]
where $K \le 8$ denotes the number of sinusoids and $\alpha_k \in \big\{ 0.200,
0.311, 0.420, 0.510, 0.630,\linebreak 0.740, 0.850, 0.970 \big\}$ their frequencies.
\end{exmp}

\begin{figure}
  \centering
  \begin{minipage}{0.48\textwidth}
    \centering
    \includegraphics[height=5cm]{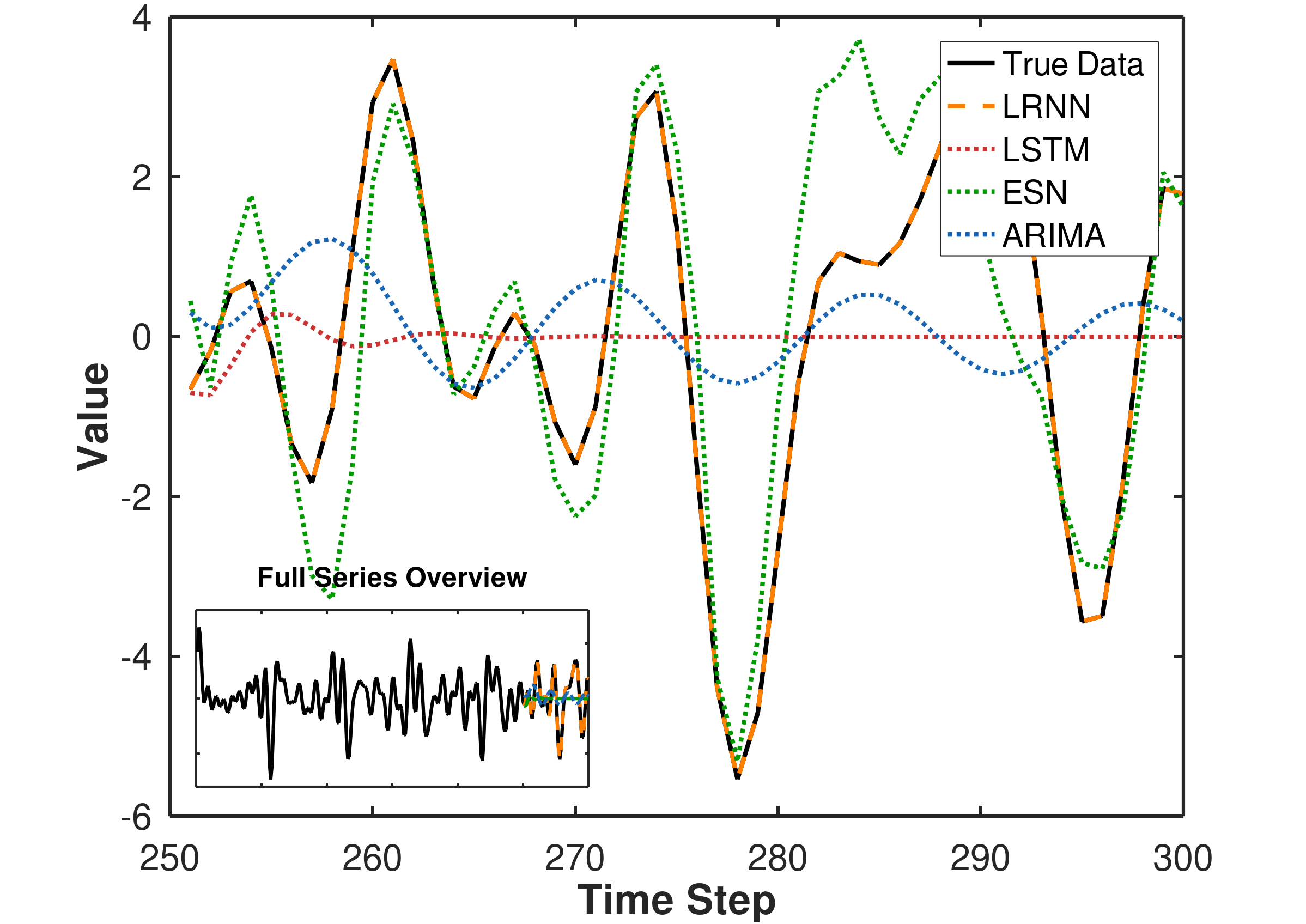}
  \end{minipage}\hfill
  \begin{minipage}{0.48\textwidth}
    \centering
    \includegraphics[height=5cm]{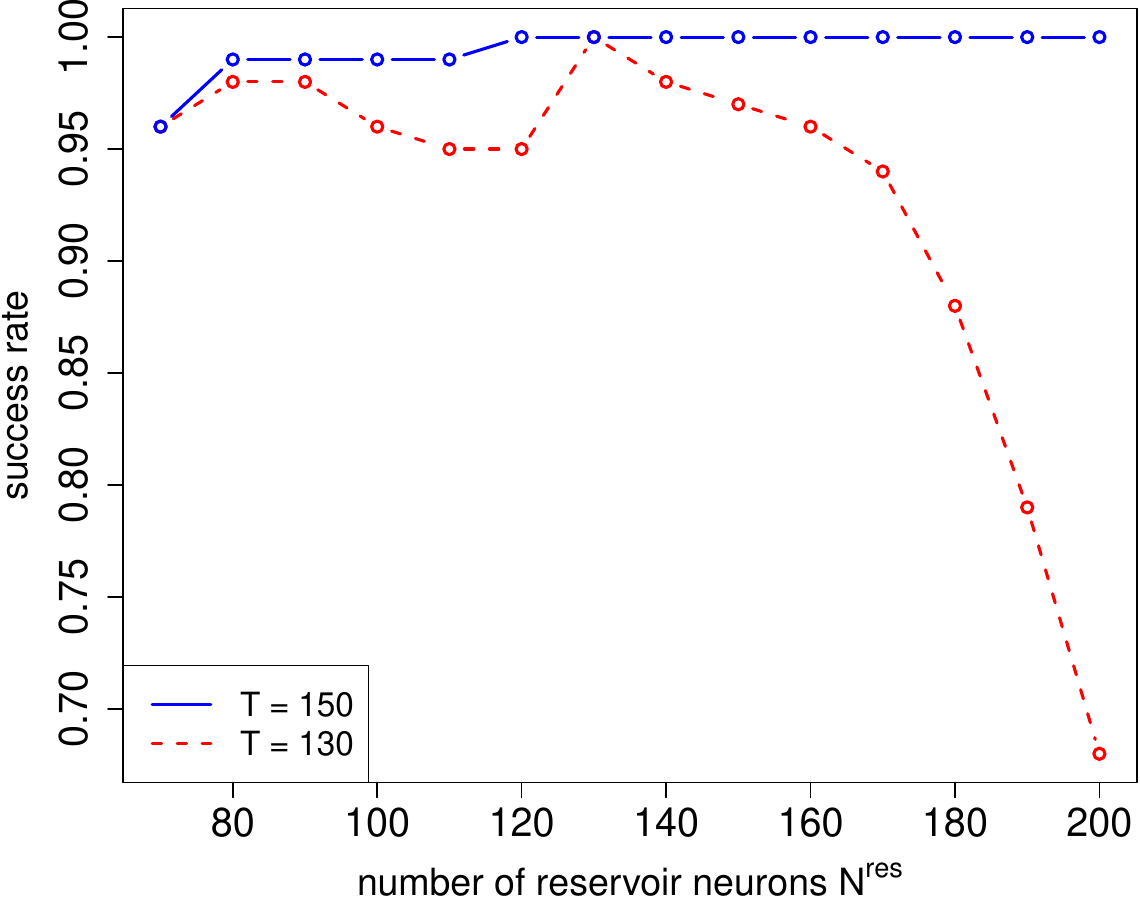}
  \end{minipage}
  \caption{\textbf{Left}: The signal $S(t)$ of $K=8$ multiple superimposed oscillators (for $1
	\le t \le 300$ and time step $\tau=1$) does not have a simple periodic
	structure (small figure). LRNN learning leads to minimal networks with
	only $N=16=2K$ reservoir neurons, i.e., two for each frequency in the
	signal, with RMSE less than $10^{-5}$ (dashed line). Other methods do
	not perform so well on the MSO benchmark (cf. dotted lines).
	\textbf{Right}: Experimental results for the MSO benchmark ($K=8$). The diagram shows the
	success rate (from 100 trials): From an initial reservoir size of $N^\mathrm{res}$
	neurons, how often is the minimal LRNN with size
	$N=16$ learned? The two curves are for different lengths $T$ of the time series
	$S(t)$ used for training. Already for $T=150$ (solid/blue), a
	minimal-size LRNN is learned in at least 96\% of the trials if
	$N^\mathrm{res} \ge 70$. For these minimal LRNNs, the RMSE is smaller
	than $10^{-5}$. As one can see, for $T=130$ (dashed/red) the given
	information does not always suffice and leads to overfitting.	}
  \label{signal}
  \label{mso8}
\end{figure}

Various publications have investigated the above MSO problem with different numbers of
sinusoids. We concentrate here solely on the most complex case $K=8$
because in contrast to other approaches it is still easy to learn for LRNNs.
Applying the LRNN learning procedure with precision threshold $\theta = 0.5$,
we arrive at LRNNs with only $N=16$ reservoir neurons and an RMSE less than
$10^{-5}$ (cf. \cref{signal}~Left), if we start with a large enough reservoir (cf.
\cref{mso8}~Right). Since two neurons are required for each frequency (cf.
\cref{ellipse}), $2K=16$ is the minimal reservoir size. Thus LRNNs outperform
the previous state-of-the-art for the
MSO task with a minimal number of units. Previous results \citep{KLB12} report $N^\mathrm{res} =
68$ as the optimal reservoir size for ESNs, but in contrast to our approach,
this number is not further reduced.

For a more systematic evaluation, we generalize the MSO benchmark (\cref{exmp})
by considering $20$ times $K=8$ random frequencies $\alpha_1,\dots,\alpha_8$
uniformly distributed on the interval $[0,1]$. We take the first $250$ time
steps\opt{long}{, i.e., $t=1,\dots,250$,} as training data and the subsequent $50$ time
steps as testing data and compare the performance of LRNNs with other approaches
(cf. \cref{related}) with respect to their RSME on the testing data, partially
using the Python library Darts \citep[see also
\url{http://unit8co.github.io/darts/}]{HLPN+22} for time series, including (a)~a
simple baseline, namely predicting constantly the arithmetic mean of the
training data, (b)~ARIMA using the AutoARIMA package \citep{HK08}, (c)~the
simple sparse ESN demo by Mantas Luko\v{s}evi\v{c}ius from
\url{http://mantas.info/code/simple_esn/} employing ReservoirPy~\citep{TPDH20},
(d)~an LSTM forecasting model implemented in Darts with hyperparameter
optimization, and (e)~LRNNs where we adopt validation data as above and choose
the best model with respect to the RMSE on the validation data from 100 trials.

\cref{sideways} shows the evaluation results. As one can see, LRNNs outperform
all other approaches on this benchmark by far. The reason for this is certainly
the network size reduction procedure (cf. \cref{reduce}), unique in our approach,
because it exactly selects the relevant components of the network: Each complex
conjugate eigenvalue pair corresponds to one of the frequencies $\alpha_1,\dots,\alpha_8$.
In general, an LRNN with $2K$ neurons suffices to represent a signal, which
might be a musical harmony \citep{Sto17b}, consisting of $K$ sinusoids (cf.
\cref{ellipse}). It can be learned by the LRNN learning procedure with network
size reduction. However, if several frequencies are close to each other (cf.
Example~\#1 in \cref{sideways}) or are rather small (cf. Example~\#12), then
LRNNs do not perform quite so well.

LRNNs are also advantageous considering the time required to train a well
performing model with less than 3\,s per series for training and testing on average:
\begin{center}
\smallskip
\begin{tabular}{cccccc}
\toprule
& Arima (R) & Arima (Python)  & ESN & LSTM & LRNN\\ \midrule
train + test time & 0.35\,s & 3.20\,s & 1.24\,s & 125.43\,s & 2.97\,s \\ \bottomrule
\end{tabular}
\smallskip
\end{center}
The ARIMA experiments with an R implementation are fastest (0.35\,s
per series) but often perform even worse than our baseline. Interestingly, an
ARIMA implementation in Python turns out to be significantly slower (3.20\,s per series).
ESN learning takes 1.24\,s per series. LRNNs more or less extend the ESN
approach and take 2.97\,s per series on average with our approach, 
including required repeated reservoir generations. LSTM learning including hyperparameter
selection takes 125.4\,s per series. All experiments are run on an Intel
i9-10940X, 3.3\,GHz CPU, and 128\,GB RAM.

%%% RESULTS (times)
% ARIMA - MSO8 on signal01..signal20: (R implementation), very fast: 0.35s / series on average (excl. file access), just the model
%            in python fast not as fast, 3.2s / series excl file access 
% ESN - MSO8 on signal00..signal20: (python implementation): 124.43s / 100 series -> 1.24s / series on average (excl file access).
% LSTM - MSO8 on signal00..20 : 125.43 s / series on average, including hyperparameter selection, excl file access
% LRNN - MSO8 on signal01..20 : 2.97 s / series on average, excl file access (averaged over 100 runs)
	
\begin{sidewaystable}
  \centering
  \begin{tabular}{c*{8}{p{8mm}}cccc@{\,+\,}cccc}
     \toprule
	\# & \multicolumn{8}{l}{frequencies} & Baseline & ARIMA & ESN & \multicolumn{3}{c}{LSTM} & \multicolumn{2}{c}{LRNN}\\
	& $\alpha_1$ & $\alpha_2$ & $\alpha_3$ & $\alpha_4$ & $\alpha_5$ & $\alpha_6$ & $\alpha_7$ & $\alpha_8$ & RMSE & RMSE & RMSE & \multicolumn{2}{c}{units} & RMSE & $N$ & RMSE\\ \midrule
	1 & 0.334 & 0.336 & 0.399 & 0.403 & 0.412 & 0.438 & 0.442 & 0.724 & 2.05613 & 2.07496 & 0.23208 & 5 & 32 & 0.16038 & 10 & \textbf{0.04761}\\
	2 & 0.049 & 0.091 & 0.161 & 0.292 & 0.472 & 0.715 & 0.832 & 0.997 & 2.25490 & 3.06428 & 0.19716 & 10 & 2 & 0.19218 & 16 & \textbf{0.00051}\\
	3 & 0.308 & 0.521 & 0.597 & 0.607 & 0.736 & 0.766 & 0.924 & 0.957 & 1.34810 & 1.46533 & 0.13680 & 5 & 8 & 0.11722 & 16 & \textbf{0.00060}\\
	4 & 0.031 & 0.348 & 0.448 & 0.476 & 0.476 & 0.613 & 0.628 & 0.833 & 2.03748 & 2.13161 & 0.25979 & 5 & 8 & 0.19209 & 14 & \textbf{0.00003}\\
	5 & 0.059 & 0.239 & 0.324 & 0.421 & 0.437 & 0.519 & 0.747 & 0.777 & 1.68045 & 2.41015 & 0.16561 & 10 & 1 & 0.14548 & 16 & \textbf{0.00011}\\
	6 & 0.013 & 0.029 & 0.262 & 0.543 & 0.636 & 0.705 & 0.740 & 0.807 & 1.79038 & 2.91774 & 0.19343 & 5 & 4 & 0.16770 & 16 & \textbf{0.00038}\\
	7 & 0.155 & 0.226 & 0.286 & 0.512 & 0.661 & 0.692 & 0.746 & 0.930 & 1.48729 & 1.75658 & 0.13109 & 5 & 4 & 0.13761 & 16 & \textbf{0.00012}\\
	8 & 0.017 & 0.027 & 0.273 & 0.475 & 0.616 & 0.848 & 0.962 & 0.989 & 1.87712 & 2.66991 & 0.22200 & 5 & 2 & 0.16481 & 16 & \textbf{0.02033}\\
	9 & 0.092 & 0.318 & 0.335 & 0.413 & 0.593 & 0.743 & 0.747 & 0.799 & 1.84272 & 2.31084 & 0.22680 & 5 & 64 & 0.16025 & 16 & \textbf{0.00142}\\
	10 & 0.108 & 0.122 & 0.262 & 0.307 & 0.391 & 0.577 & 0.589 & 0.603 & 1.64237 & 1.66822 & 0.18361 & 10 & 16 & 0.13289 & 16 & \textbf{0.00772}\\
	11 & 0.071 & 0.264 & 0.557 & 0.609 & 0.641 & 0.719 & 0.853 & 0.964 & 1.71070 & 2.01792 & 0.13613 & 5 & 16 & 0.14964 & 16 & \textbf{0.00003}\\
	12 & 0.036 & 0.052 & 0.062 & 0.222 & 0.279 & 0.316 & 0.563 & 0.672 & 1.52224 & 1.83523 & 0.18562 & 10 & 8 & \textbf{0.15571} & 14 & 0.15984\\
	13 & 0.036 & 0.481 & 0.571 & 0.724 & 0.750 & 0.750 & 0.864 & 0.898 & 2.51888 & 2.50217 & 0.15961 & 5 & 16 & 0.22408 & 14 & \textbf{0.00067}\\
	14 & 0.175 & 0.220 & 0.258 & 0.419 & 0.487 & 0.513 & 0.628 & 0.663 & 2.17787 & 1.90370 & 0.23698 & 5 & 8 & 0.19731 & 16 & \textbf{0.00069}\\
	15 & 0.185 & 0.300 & 0.461 & 0.751 & 0.814 & 0.833 & 0.840 & 0.992 & 2.18527 & 1.63141 & 0.15527 & 5 & 2 & 0.20659 & 14 & \textbf{0.03709}\\
	16 & 0.088 & 0.120 & 0.137 & 0.245 & 0.478 & 0.793 & 0.797 & 0.992 & 2.27108 & 2.51581 & 0.15872 & 10 & 1 & 0.19325 & 16 & \textbf{0.01439}\\
	17 & 0.002 & 0.242 & 0.348 & 0.503 & 0.734 & 0.748 & 0.759 & 0.862 & 1.47542 & 1.76223 & 0.23496 & 10 & 8 & 0.13808 & 16 & \textbf{0.00150}\\
	18 & 0.018 & 0.352 & 0.583 & 0.625 & 0.714 & 0.824 & 0.838 & 0.888 & 2.44582 & 2.71492 & 0.24585 & 5 & 32 & 0.25433 & 16 & \textbf{0.00010}\\
	19 & 0.046 & 0.105 & 0.263 & 0.351 & 0.517 & 0.556 & 0.758 & 0.807 & 1.79806 & 2.24052 & 0.18220 & 5 & 16 & 0.18124 & 16 & \textbf{0.00005}\\
	20 & 0.091 & 0.141 & 0.375 & 0.578 & 0.686 & 0.785 & 0.951 & 0.996 & 2.05839 & 2.21768 & 0.15490 & 5 & 32 & 0.17551 & 16 & \textbf{0.00001}\\
	\bottomrule
  \end{tabular}
\caption{Evaluation results for $20$ generalized MSO examples with $8$ randomly
generated frequencies each. The best performing approach is highlighted by bold
face. For the LSTMs, the number of input and hidden units of the best performing
neural network is given. For LRNNs, the network size $N$ after network size
reduction is shown.}
\label{sideways}
\end{sidewaystable}

\subsection{Solving Number Puzzles}

\begin{exmp}
Number series tests are a popular type of intelligence test. The function
represented by a number series can often be learned also by artificial neural
networks, in particular RNNs. In \citep{GW13}, 20 number puzzles are listed \citep[cf.][]{RK11}.
Among them are the series:

\[ \begin{array}{c@{\,=\,}lc@{\,=\,}l}
	S_8 & [28,33,31,36,34,39,37,42] & f(t) & f(t-2) + 3\\[3pt]
	S_9 & [3,6,12,24,48,96,192,384] & f(t) & 2 f(t-1)\\[3pt]
	S_{15} & [6,9,18,21,42,45,90,93] & f(t) & 2 f(t-2) + 4.5 + 1.5 (-1)^{t-1}\\[3pt]
	S_{19} & [8,12,16,20,24,28,32,36] & f(t) & f(t-1) + 4
\end{array} \]
\end{exmp}

We apply the LRNN learning procedure to all 20 number puzzles taking small
reservoirs because the number series are short. As a side effect, this leads to
learning more general functions, which seems to be fully adequate because number
puzzles are usually presented to humans. The first 7 of 8 elements of each
series are given as input. In each trial, we repeatedly generate LRNNs, until the
RMSE is smaller than $\theta=0.1$. Then the last ($8^\text{th}$) element of the
respective series is predicted (according to \cref{predict}) and rounded to the
nearest integer because all considered number series are integer.

\cref{table} lists the percentages of correct predictions of the last
element for different settings. Here, the series with definitions recurring to
$f(t-2)$ but not $f(t-1)$, e.g., $S_8$ and $S_{15}$, turned out to be the most
difficult. If we add the previous values of the time series, i.e., $f(t-2)$,
as clue to the input, then the correctness of the procedure increases
significantly: For 19 of 20 number puzzles, the most frequently predicted last
element (simple majority) is the correct one. It is predicted in 76.5\% on
average over all trials and number puzzles. Let us remark that the whole
evaluation with altogether $20 \cdot 50 \cdot 1000 = 1\,000\,000$ trials
including possibly repeated network generation ran in only a few minutes on
standard hardware.

Although the application domains of transformer models \cite{VS+17} and LRNNs are different,
one can compare modern LLMs and LRNNs with respect to solving number puzzles:
LLMs can yield interpretable explanations. For example, ChatGPT (which is based on
GPT-4, cf. \cref{regress}) correctly predicts the next value for all number
puzzles, but in 2 of 20 cases finds incorrect (recursive or closed) formulae explaining the number puzzles\opt{proof}{ (cf. \cref{chatgpt})}.
The LRNN learning procedure with network size reduction determines LRNNs with
small reservoir sizes. From these, we can read off explicit, i.e., also
interpretable, small formulae for the number series, e.g., $f(t) = 3 \cdot 2^n$
for $S_9$ with a minimal reservoir size $N^\mathrm{res}=1$. Focusing on small
reservoir sizes (determined from $1000$ trials), for 18 of 20 number puzzles,
the most frequently predicted last element is the correct one.
Hence, network size reduction can improve the generalization and explanation
ability of LRNNs because it may suppress noise. However, for complex applications with
high variability and noise, the network size reduction procedure clearly does
not lead to such a strong compression of the network (cf. \cref{soccer}).

\begin{table}[h]
  \centering
  \sisetup{table-number-alignment = center, %
	table-figures-integer = 3, table-figures-decimal = 1, table-figures-exponent = 0}%
  \begin{tabular}{crrrrcr}
	\toprule
 	\head{series} & \head{$N^\mathrm{res}=3$}& \head{$N^\mathrm{res}=4$} & \head{$N^\mathrm{res}=5$} & \head{with reduction} &  \head{$\min(N^\mathrm{res})$} & \head{plus clue}\\ \midrule
	$S_{1}$ & 2.2\% & 1.3\% & 1.3\% & 64.4\% & 3 & 33.4\% \\
	$S_{2}$ & 37.6\% & 42.2\% & 29.4\% & 100.0\% & 2 & 100.0\% \\
	$S_{3}$ & 5.4\% & 4.1\% & 1.1\% & 99.5\% & 2 & 100.0\% \\
	$S_{4}$ & 23.8\% & 24.2\% & 16.8\% & 81.5\% & 2 & 99.9\% \\
	$S_{5}$ & 56.9\% & 57.6\% & 44.2\% & 99.1\% & 2 & 99.7\% \\
	$S_{6}$ & 31.7\% & 33.7\% & 16.1\% & 56.6\% & 2 & 100.0\% \\
	$S_{7}$ & 72.8\% & 68.2\% & 56.2\% & 99.2\% & 2 & 100.0\% \\
	$S_{8}$ & 5.1\% & 3.4\% & 1.3\% & 86.0\% & 3 & 76.3\% \\
	$S_{9}$ & 100.0\% & 100.0\% & 100.0\% & 100.0\% & 1 & 100.0\% \\
	$S_{10}$ & 48.9\% & 71.5\% & 67.6\% & 83.3\% & 2 & 100.0\% \\
	$S_{11}$ & 10.6\% & 9.0\% & 3.4\% & 96.9\% & 2 & 100.0\% \\
	$S_{12}$ & 23.8\% & 21.1\% & 11.0\% & 82.4\% & 2 & 43.2\% \\
	$S_{13}$ & 56.5\% & 58.1\% & 41.5\% & 95.1\% & 2 & 99.8\% \\
	$S_{14}$ & 6.7\% & 7.4\% & 2.1\% & 94.3\% & 2 & 87.1\% \\
	$S_{15}$ & 1.6\% & 2.6\% & 2.5\% & 3.6\% & 4 & 1.1\% \\
	$S_{16}$ & 6.8\% & 5.9\% & 3.4\% & 88.7\% & 3 & 73.3\% \\
	$S_{17}$ & 11.9\% & 12.0\% & 6.8\% & 51.6\% & 3 & 41.0\% \\
	$S_{18}$ & 3.1\% & 2.0\% & 1.1\% & 37.5\% & 4 & 18.0\% \\
	$S_{19}$ & 59.6\% & 70.1\% & 72.0\% & 99.0\% & 2 & 99.8\% \\
	$S_{20}$ & 1.5\% & 0.5\% & 0.6\% & 57.9\% & 4 & 57.2\% \\ \bottomrule
  \end{tabular}
  \caption{Percentages of correct predictions of the last element for 20 number
	puzzles \citep{RK11,GW13} in 1000 trials for different settings:
	\textbf{(a)}~with fixed reservoir size $N^\mathrm{res}=3,4,5$;
	\textbf{(b)}~with network size reduction starting with
	$N^\mathrm{res}=7$ reservoir neurons yielding small/minimal reservoirs;
	\textbf{(c)}~same procedure but in addition always the previous time
	series value is used as clue.}
  \label{table}
\end{table}

\subsection{Replaying Soccer Games}\label{soccer}

RoboCup \citep{KA+97} is an international scientific robot competition in which
teams of multiple robots compete against each other. Its different leagues
provide many sources of robotics data that can be used for further analysis and
application of machine learning. A \emph{soccer simulation} game lasts 10\,mins and is
divided into 6000 time steps where the length of each cycle is 100\,ms. Logfiles
contain information about the game, in particular about the current positions of
all players and the ball including velocity and orientation for each cycle.
For our experiments we evaluated ten games of the top-five teams in a public
research dataset \citep{MO+19} that contains recorded games from the RoboCup 2D
soccer simulation league \citep{CDF+03,GFG16} in 2016 and 2017
\citep{MO+18}\opt{unblind}{ (dataset available from
\url{http://bitbucket.org/oliverobst/robocupsimdata})}. We consider the
$(x,y)$-coordinates of the ball and all 22 players for all time points during
regular game play, the so-called \emph{play-on} mode.

\begin{table}
\centering
\sisetup{table-number-alignment = center,  % group-digits = false,
	      table-figures-integer = 1, table-figures-decimal = 5, table-figures-exponent = 0}%
\begin{tabular}[b]{cSScc}
	\toprule
	\head{game} & \head{RMSE\,(1)} & \head{RMSE\,(2)} & \head{$N$} & \head{reduction}\\ \midrule
	\#1 & 0.00000 & 0.78581 & 385 & 29.5\%\\
	\#2 & 0.00001 & 0.96911 & 403 & 26.2\% \\
	\#3 & 0.00004 & 0.97680 & 390 & 28.6\% \\
	\#4 & 0.00000 & 0.97696 & 406 & 25.6\% \\
	\#5 & 0.00000 & 0.98425 & 437 & 20.0\% \\
	\#6 & 0.00000 & 0.47939 & 354 & 35.2\% \\
	\#7 & 0.00000 & 0.78731 & 390 & 28.6\% \\
	\#8 & 0.00255 & 0.98787 & 385 & 29.5\% \\
	\#9 & 0.00000 & 0.88518 & 342 & 37.4\% \\
	\#10 & 0.00000 & 0.96043 & 376 & 31.1\% \\
	\bottomrule
  \end{tabular}
  \caption{For ten RoboCup simulation games, an LRNN is learned with initially
	$N = 500+46 = 546$ neurons. The table shows the RMSE
	(1)~before and (2)~after dimensionality reduction where $\theta=1$\,m. The
	network size can be reduced significantly -- 29.2\% on average
	(last column).}
  \label{robocup}
\end{table}

\begin{figure}
\centering
\includegraphics[width=0.54\columnwidth]{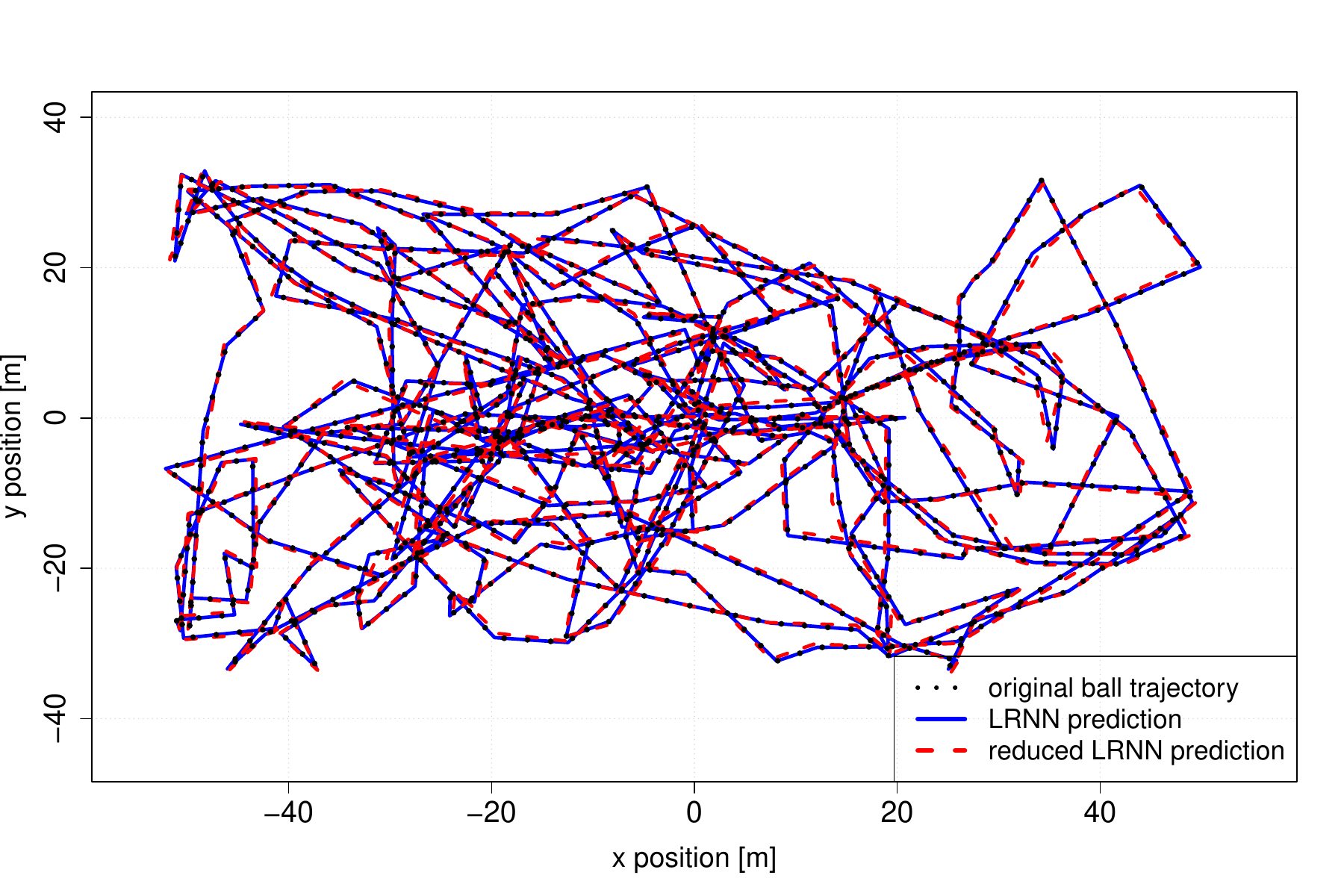} % 30.48x20.32
\hfill %
%\includegraphics[width=0.2\columnwidth]{fig/goalie}     % 3450x4373. (23)
%\hfill %
\includegraphics[width=0.43\columnwidth]{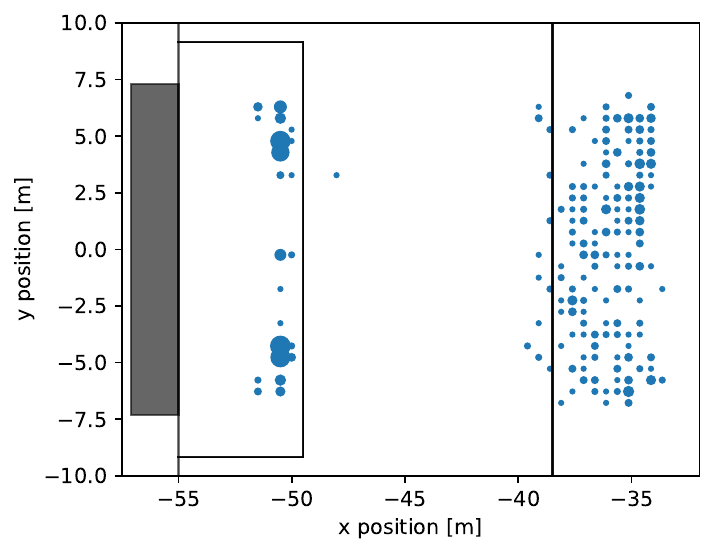} % 5717x4437 (38)
\caption{\textbf{Left:} Ball trajectory of RoboCup 2D soccer simulation game \#6
	(\emph{Oxsy}~0 versus \emph{Gliders}~2016) on a pitch of size
	$105\,\mathrm{m} \times 68\,\mathrm{m}$. For all 6000 time steps, the
	original trajectory of the ball during play is shown (dotted/black). The
	game can be replayed by an LRNN with $N = 500+46 = 546$ neurons with high
	accuracy (solid/blue). The reduced network with $N = 354$ reservoir
	neurons still mimics the trajectory with only small error (dashed/red).
%Middle: Trajectory of the FRA-UNIted goalkeeper in front of the goal
%during games at the RoboCup Japan Open 2020. 
\textbf{Right:} Dots (in blue) mark positions
that were visited more than three times (larger dots: more visits, 0.5\,m resolution),
information that can be derived from predictions, highlighting three larger,
frequently visited regions in front of the goal.}
\label{game}
\end{figure}

For LRNN learning, we use only every $10^\text{th}$ time step of each 6000 step game with $d =
2+2 \cdot 22 = 46$ input dimensions and start with a reservoir consisting of
$N^\mathrm{res} = 500$ neurons. We repeat the learning procedure until the RMSE
is smaller than $1$; on average, already two attempts suffice for this. This
means, if we replay the game by the learned LRNN (in output generating mode)
then on average the predicted positions deviate less than 1\,m from the real
ones (Euclidean distance) -- over the whole length of the game (cf.
\cref{game}~Left). Network size reduction leads to significantly less neurons
compared to the original number $N = 46+500 = 546$ -- on average 29.2\% if we
concentrate on the relevant components for the ball trajectory (cf.
\cref{robocup}). Note that the size of the learned network is in $O(N)$ (cf.
\cref{complexity}). Thus, the LRNN model is smaller than the original time
series representation of a game, which means it is substantially compressed. The
complete learning procedure runs in less than a minute on standard hardware.

\cref{multi} shows how we can learn from multiple time series at once. This is
also helpful here because by this procedure we can investigate the overall
behavior of a specific robot soccer agent. As example for this, we consider the
trajectories of the goalkeeper of the RoboCup simulation team FRA-UNIted during
the seeding and the qualifying round of RoboCup Japan Open 2020 (see
\url{http://bit.ly/japanopen2020ssim}). For learning one LRNN from this, we employ a
reservoir with $N^\mathrm{res} = 1000$ neurons, adopt again a maximum threshold
for the RMSE of $\theta = 1$\,m, and only use every $20^\text{th}$ step of each of the
$7$ games. The overall trajectory of the FRA-UNIted goalkeeper can be learned
easily then (cf. \cref{game}~Right). From this, one may conclude that the
goalkeeper takes up three basic positions in front of the goal, does not
approach the center line more than about 30\,m and hardly leaves the center
line.

\subsection{Predicting Stock Prices}\label{stock}

Stock price prediction is a topic that receives a considerable amount of
attention in finance. Complexity of markets resulting in multiple and sudden
changes in trends of stock prices make their prediction a challenge.
Consequently, a number of different approaches and methods have been developed.
\citep{Lit20} analyzes $30$ different stocks by ARIMA and LRNNs using the
closing stock prices 2016--2019 from \url{http://de.finance.yahoo.com/}. The stock price time series (consisting of
$762$ data points each) are split into training and testing data, with the first
80\% of each series for training and the final 20\% for evaluation. For a
representative comparison, the RMSE of the predictions on every stock in the set
is calculated. The average RMSE using LRNNs with $N^\mathrm{res}=600$ reservoir
neurons is $E_\mathrm{test}=18.40${\,\euro}, lower than the average RMSE using
ARIMA models with seasonal patterns modeled using Fourier terms
\citep[p.~321]{HA13} which is $E_\mathrm{test}=24.23${\,\euro}. For shorter term
predictions of 60 steps, it is possible to slightly reduce the RMSE further to
E$_\mathrm{test}=17.46${\,\euro} by using smaller LRNNs of $N^\mathrm{res}=200$
reservoir neurons and a smaller training set of $240$ data points. With an
average stock price of $286.71${\,\euro} of all stocks in the set, the average
deviation is only~6.1\%.

Apart from the good prediction results, LRNNs have the advantage that they allow
the prediction of multiple stocks at the same time. An LRNN can read in 30
stocks and predict each of them concurrently. For a concurrent forecast for $60$
steps, LRNNs achieve an average RMSE of
$E_\mathrm{test}=30.02${\,\euro} with $240$ training steps. Compared to ARIMA,
LRNNs have also an advantage when it comes to the number of hyperparameters that
have to be tuned. The LRNN model is robust when it comes to choosing the number
of reservoir neurons, whereas the ARIMA model requires the adjustment of many
parameters (e.g., for seasonal patterns). The compute time for ARIMA increases
significantly with the number of hyperparameters. For the considered $30$
stocks, LRNNs are computed about 15 times faster than the ARIMA models with the
selected number of Fourier terms.

For a more systematic evaluation, we take the stocks of the German stock market
index DAX (again from \url{http://de.finance.yahoo.com/}) and consider the last
$250+50$ data points until the end of 2021 as training and testing data,
respectively, for each stock in the DAX existing at least $300$ trading days
until that time. We apply the same approaches as in \cref{mso} and compute their
RMSE with respect to the testing data, each normalized\opt{long}{ (i.e., divided)}
by the arithmetic mean of the corresponding training data. As baseline we just
constantly predict the stock price of the last trading day of the training data.

\begin{figure}
  \centering
  \includegraphics[width=0.5\textwidth]{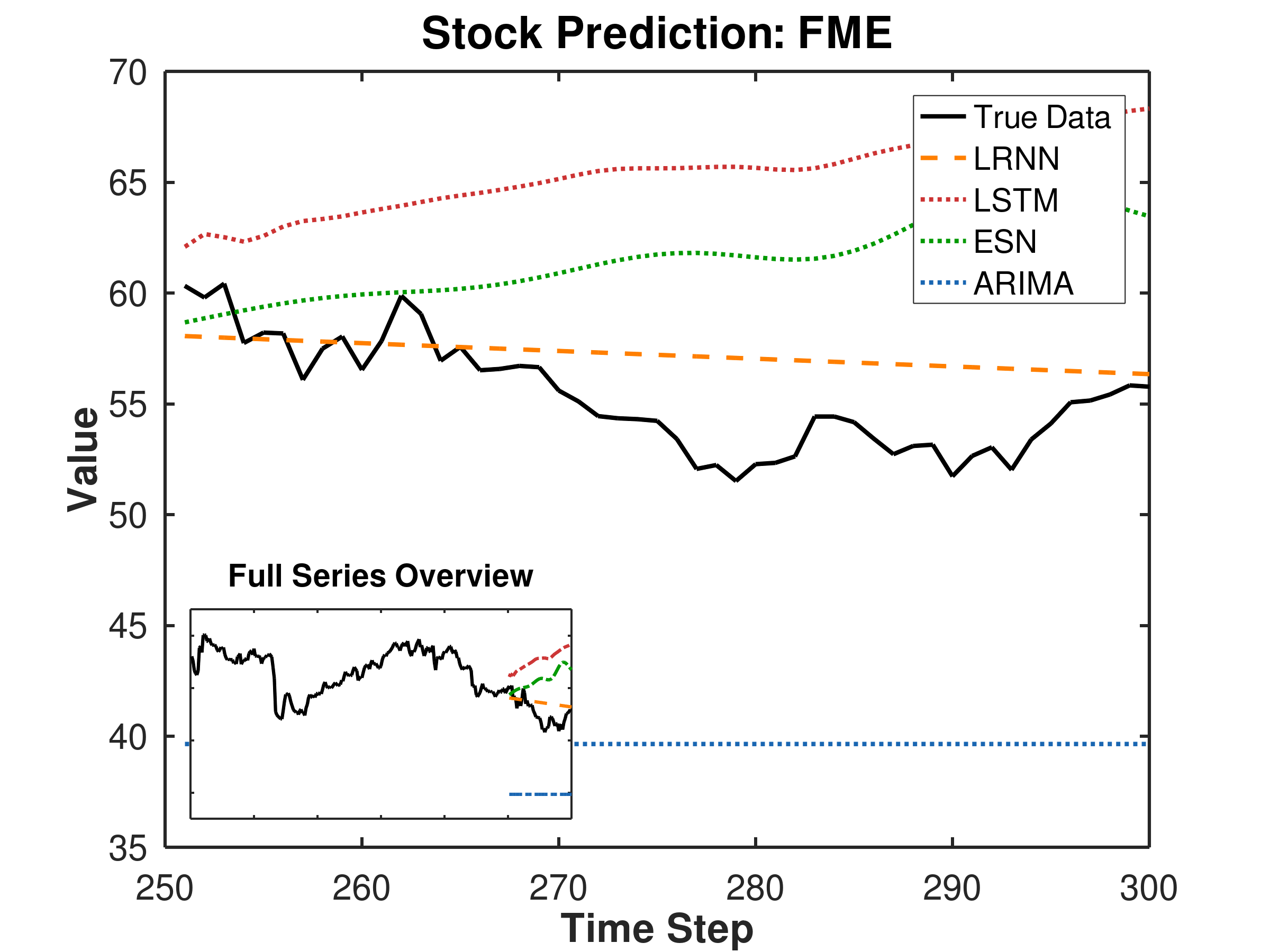}
  \caption{Stock price prediction for Fresenius Medical Care (FME.DE) with LRNNs
	(dashed line) and other approaches. The overall performance for all
	approaches often is not much better than the baseline (cf. \cref{dax}).}
  \label{stockfig}
\end{figure}

As one can see in \cref{dax}, LRNNs outperform the other approaches in the
majority of all cases, namely for 19~of~39 stocks, where the resulting network
size $N$ after size reduction often is rather small. However, the overall
performance for all approaches often is not much better than the baseline (see
also \cref{stockfig}). In line with that, the reviews in
\citep{SIZ19,SGO20,Jia21} show that predicting stock prices remains a
challenging problem, especially for a longer timeframe, which we investigate
here. Contemporary research often uses complex models, ranging from LSTM RNNs
\citep{NPO17,RPV17} to attention-based models with transformer architecture
\citep{VS+17} that use further information about the events that drive the stock
prices, e.g., news texts from social media \citep{LL+19}. These models and also
LRNNs yield rather accurate results, but mainly only in the short run.

\begin{table}[b]
\begin{tabular}{l@{ }crccccc}
\toprule
\multicolumn{2}{c}{DAX member} & Baseline & ARIMA & ESN & LSTM & \multicolumn{2}{l}{LRNN}\\
name & stock & RMSE & RMSE & RMSE & RMSE & $N$ & RMSE\\ \midrule
Adidas & ADS.DE & \textbf{0.0554} & 0.0554 & 0.0784 & 0.1162 & 2 & 0.0569\\
Airbus & AIR.DE & 0.0618 & 0.1070 & 0.0583 & 0.2293 & 2 & \textbf{0.0500}\\
Allianz & ALV.DE & 0.0252 & 0.0241 & \textbf{0.0199} & 0.0916 & 2 & 0.0503\\
BASF & BAS.DE & 0.0438 & \textbf{0.0262} & 0.0739 & 0.1218 & 2 & 0.0433\\
Bayer & BAYN.DE & 0.0414 & 0.0409 & 0.0626 & 0.1027 & 1 & \textbf{0.0350}\\
BMW St & BMW.DE & 0.0706 & \textbf{0.0706} & 0.0744 & 0.1509 & 11 & 0.0787\\
Brenntag & BNR.DE & \textbf{0.0544} & 0.0938 & 0.0664 & 0.1162 & 2 & 0.0969\\
Continental & CON.DE & 0.0518 & \textbf{0.0518} & 0.0803 & 0.2540 & 2 & 0.1918\\
Covestro & COV1.DE & \textbf{0.0488} & 0.0488 & 0.0564 & 0.1612 & 1 & 0.0609\\
Deutsche Börse & DB1.DE & 0.0307 & 0.0307 & \textbf{0.0249} & 0.0814 & 1 & 0.0347\\
Deutsche Bank & DBK.DE & 0.0442 & 0.0442 & \textbf{0.0229} & 0.1794 & 1 & 0.0599\\
Delivery Hero & DHER.DE & 0.1119 & 0.1040 & 0.1148 & 0.2226 & 3 & \textbf{0.0798}\\
Deutsche Post & DPW.DE & 0.0487 & 0.0379 & \textbf{0.0369} & 0.1822 & 1 & 0.0549\\
Deutsche Telekom & DTE.DE & 0.0279 & 0.0279 & 0.0619 & 0.1209 & 2 & \textbf{0.0224}\\
Deutsche Wohnen SE & DWNI.DE & 0.2410 & 0.2410 & 0.2468 & \textbf{0.1706} & 1 & 0.2589\\
Siemens Energy & ENR.DE & 0.0457 & 0.0457 & 0.0364 & 0.2212 & 9 & \textbf{0.0349}\\
E.ON & EOAN.DE & 0.0646 & 0.0646 & 0.0666 & 0.4041 & 1 & \textbf{0.0389}\\
Fresenius Medical Care & FME.DE & 0.0831 & 0.0842 & 0.0916 & 0.1279 & 1 & \textbf{0.0738}\\
Fresenius & FRE.DE & 0.1256 & 0.1256 & 0.1560 & 0.1399 & 2 & \textbf{0.1228}\\
HeidelbergCement & HEI.DE & 0.0504 & \textbf{0.0259} & 0.0708 & 0.1171 & 2 & 0.0912\\
Henkel Vz & HEN3.DE & 0.0476 & 0.0476 & 0.0328 & 0.1479 & 2 & \textbf{0.0237}\\
HelloFresh & HFG.DE & 0.1236 & 0.1236 & 0.1241 & 0.3065 & 2 & \textbf{0.1220}\\
Infineon & IFX.DE & 0.1035 & 0.1029 & 0.1623 & 0.3574 & 1 & \textbf{0.0705}\\
Linde PLC & LIN.DE & 0.1077 & \textbf{0.0710} & 0.1134 & 0.1803 & 1 & 0.0909\\
Merck KGaA & MRK.DE & 0.1293 & 0.0826 & \textbf{0.0332} & 0.2156 & 1 & 0.0361\\
MTU Aero Engines & MTX.DE & \textbf{0.0598} & 0.0600 & 0.1085 & 0.2046 & 2 & 0.0652\\
Münchener Rück & MUV2.DE & 0.0257 & \textbf{0.0256} & 0.0430 & 0.1276 & 1 & 0.0421\\
Porsche Vz & PAH3.DE & \textbf{0.0690} & 0.1116 & 0.0920 & 0.1710 & 2 & 0.1188\\
Puma & PUM.DE & 0.0894 & 0.0887 & 0.0783 & 0.1737 & 1 & \textbf{0.0531}\\
Qiagen & QIA.DE & 0.0660 & 0.0660 & 0.0810 & 0.1311 & 2 & \textbf{0.0225}\\
RWE & RWE.DE & 0.0460 & 0.0460 & 0.0805 & 0.1202 & 2 & \textbf{0.0276}\\
SAP & SAP.DE & 0.0398 & \textbf{0.0398} & 0.0459 & 0.1200 & 2 & 0.0442\\
Siemens Healthineers & SHL.DE & 0.1057 & 0.0561 & 0.0975 & 0.1898 & 1 & \textbf{0.0328}\\
Siemens & SIE.DE & 0.0561 & \textbf{0.0354} & 0.0758 & 0.1276 & 2 & 0.0996\\
Sartorius Vz & SRT3.DE & 0.0676 & 0.0676 & 0.0591 & 0.2400 & 1 & \textbf{0.0463}\\
Symrise & SY1.DE & 0.1032 & 0.1032 & 0.0893 & 0.1055 & 1 & \textbf{0.0755}\\
Vonovia & VNA.DE & 0.0714 & 0.0801 & 0.0662 & 0.1222 & 1 & \textbf{0.0593}\\
Volkswagen Vz & VOW3.DE & \textbf{0.0615} & 0.0615 & 0.1207 & 0.1521 & 2 & 0.1181\\
Zalando & ZAL.DE & 0.0656 & 0.0656 & 0.1139 & 0.2111 & 2 & \textbf{0.0600}\\
\bottomrule
\end{tabular}
\caption{Evaluation results for all stocks in the DAX existing at least 300
trading days until the end of 2021. The best performing
approach is highlighted by bold face. For LRNNs, the network size $N$ after size
reduction is shown.}
\label{dax}
\end{table}

\section{Conclusions}\label{conclude}

This paper presented LRNNs as an efficient and mathematically grounded approach
to time series modeling. The main contribution is a closed-form learning
procedure that requires solving only a linear equation system and eliminates the
need for backpropagation or gradient descent. This enables fast and stable
training while retaining strong approximation capabilities. LRNNs, despite using
purely linear activation functions, can represent a wide class of functions,
including exponential, polynomial, and trigonometric signals. A second key
contribution is the method for learning network architecture through spectral
analysis of the transition matrix. This enables substantial and principled
reduction in network size by identifying the most relevant dynamical modes based
on eigenvalue contributions. Unlike iterative pruning, our approach reduces the
network in a single step and often yields compact models with interpretable
internal structure.

We have demonstrated this across several tasks, including the MSO benchmark,
stock prediction, and number puzzles, where LRNNs matched or outperformed LSTMs
and ESNs with significantly fewer units and faster runtime. The models also
showed robustness in nonstationary and irregular data, though very large
reservoirs may cause numerical issues, especially in long-term prediction.
Nonetheless, the reduced models typically generalize well and support compact
formula extraction for interpretable applications. These properties make LRNNs
particularly suitable for domains requiring fast training, compact models, and
insights into learned structure, for example, finance, embedded systems, and
symbolic reasoning. Future work may explore hybrid extensions that incorporate
nonlinearity or adaptive updates, while preserving the architectural simplicity
and closed-form solvability of the LRNN framework.

\opt{unblind}{%
\section*{Acknowledgments}

We would like to thank Chad Clark, Andrew Francis, Rouven Neitzel, Oliver Otto,
Kai Steckhan, Flora Stolzenburg, and Ruben Zilibowitz, as well as several
anonymous referees for helpful discussions
and comments. The research reported in this paper has been supported by the
German Academic Exchange Service (DAAD) by funds of the German Federal Ministry
of Education and Research (BMBF) in the Programmes for Project-Related Personal
Exchange (PPP) under grant no.~57319564 and Universities Australia (UA) in the
Australia-Germany Joint Research Cooperation Scheme within the project
\emph{\underline{De}ep \underline{Co}nceptors for Tempo\underline{r}al
D\underline{at}a M\underline{in}in\underline{g}} (Decorating).
A first short and preliminary version of this paper was presented at the conference
\emph{Cognitive Computing} in Hannover \citep{SMO18}. It received the prize for the
most technologically feasible poster contribution.
}

\bibliography{recupred}

\opt{proof}{\newpage\appendix

\section{Proof of \cref{rem}}\label{proof:rem}

\noindent For a function $x$ and its first derivative $\dot{x}$ with
respect to time $t$, we have
\[
  \dot{x}(t) = \lim\limits_{\tau \to 0} \frac{x(t+\tau)-x(t)}{\tau}
	\quad\text{and hence}\quad
  x(t+\tau) \approx x(t) + \tau\,\dot{x}(t)
\]
for small time steps $\tau>0$. We can apply this to $x^{(k)}(t)$ for $k \ge 0$
in the difference of \cref{diff} between the times $t+\tau$ and $t$ divided by
$\tau$ and obtain:
\begin{eqnarray*}
  0 & = & \frac{\displaystyle\sum_{k=0}^n c_k\,x^{(k)}(t+\tau)-\sum_{k=0}^n c_k\,x^{(k)}(t)}{\tau}\\
    & = & \sum_{k=0}^{n-1} c_k \frac{x^{(k)}(t+\tau)-x^{(k)}(t)}{\tau} + c_n \frac{x^{(n)}(t+\tau)-x^{(n)}(t)}{\tau}\\
    & \approx & \sum_{k=0}^{n-1} c_k\,x^{(k+1)}(t) + \frac{c_n}{\tau} \left(x^{(n)}(t+\tau)-x^{(n)}(t)\right)
\end{eqnarray*}
This is equivalent to:
\[
	x^{(n)}(t+\tau)
	\approx x^{(n)}(t) - \frac{\tau}{c_n}\,\sum_{k=0}^{n-1} c_k\,x^{(k+1)}(t)
\]
From this, we can read off the desired transition matrix $W$ of size $(n+1) \times (n+1)$ from \cref{mat}.
Together with a start vector $s$ satisfying \cref{diff}, we can thus solve
differential equations approximately by LRNNs.

\section{Real Jordan Canonical Form}\label{omi}

For a completely real-valued decomposition \citep[Section~3.4.1]{HJ13}, a Jordan
matrix $J$ can be transformed as follows:
\begin{enumerate}
  \item A Jordan block with real eigenvalue $\lambda$ remains as is in $J$.
  \item For complex conjugate eigenvalue pairs $\lambda = \lambda_\Re +
	\mathfrak{i}\lambda_\Im$ and $\overline{\lambda} = \lambda_\Re - \mathfrak{i}\lambda_\Im$
	with $\lambda_\Re,\lambda_\Im \in \mathbb{R}$, the
	direct sum of the two corresponding Jordan blocks $J_m(\lambda)$ and
	$J_m(\overline{\lambda})$ is replaced by one real Jordan block:
	\[
	  \left[ \begin{array}{*{5}{c}}
		M & I & O & \cdots & O\\
		O & M & I & \ddots & \vdots\\
		\vdots & \ddots & \ddots & \ddots & O\\
		\vdots & & \ddots & M & I\\
		O & \cdots & \cdots & O & M
	  \end{array} \right]
	\]
	with
	  $M = \left[ \begin{array}{cc}
		\lambda_\Re & \lambda_\Im\\
		-\lambda_\Im & \lambda_\Re
	  \end{array} \right]$,
	  $I = \left[ \begin{array}{cc}
		1 & 0\\
		0 & 1
	  \end{array} \right]$, and
	  $O = \left[ \begin{array}{cc}
		0 & 0\\
		0 & 0
	  \end{array} \right]$.
\end{enumerate}
\smallskip

This procedure yields a real Jordan matrix $J$. In consequence, we have to
transform $V$ also into a completely real-valued form. For this, for each complex
conjugate eigenvalue pair $\lambda$ and $\overline{\lambda}$, the corresponding
eigenvectors in $V$ can be replaced by real-valued vectors.

\section{Proof of \cref{chad}}\label{proof:chad}

We first prove the case where the Jordan matrix $J$ only contains ordinary
Jordan blocks as in \cref{jordan}, i.e., possibly with complex eigenvalues
on the diagonal. Since $J$ is a direct sum of Jordan blocks, it suffices to
consider the case where $J$ is a single Jordan block because, as the Jordan
matrix $J$, the matrices $A$ and also $B$ (see below) can be obtained as direct
sums, too.

In the following, we use the column vectors $y = \big[ y_1 \cdots y_N
\big]^\top$ with all nonzero entries, $x = \big[ x_1 \cdots x_N \big]^\top$
with $x = V^{-1} \cdot s$ (cf. \cref{eigen}), and $b = \big[ b_1 \cdots b_N
\big]^\top$. From $b$, we construct the following upper triangular Toeplitz
matrix
\[ B = \left[ \begin{array}{*{4}{c}}
  b_N & \cdots & b_2 & b_1\\
  0 & \ddots & & b_2\\
  \vdots & \ddots & \ddots & \vdots\\
  0 & \cdots & 0 & b_N
\end{array} \right] \]
which commutes with the Jordan block $J$ \citep[Section~3.2.4]{HJ13}, i.e., it
holds that (a)~$J \cdot B = B \cdot J$. We define $B$ and hence $b$ by the
equation (b)~$x = B \cdot y$ which is equivalent to:

\[ \left[ \begin{array}{*{4}{c}}
  y_N & \cdots & y_2 & y_1\\
  0 & \ddots & & y_2\\
  \vdots & \ddots & \ddots & \vdots\\
  0 & \cdots & 0 & y_N
\end{array} \right] \cdot b = \left[ \begin{array}{c}
  x_1\\[7pt] x_2\\ \vdots\\ x_N
\end{array} \right] \]
Since the main diagonal of the left matrix contains no $0$s because $y_N \neq 0$
by precondition, there always exists a solution for $b$ \citep[Section~0.9.3]{HJ13}.
Then $A = V \cdot B$ does the job:
\[ f(t) = W^t \cdot s
	\overset{\text{\cref{jordan}}}{=} V \cdot J^t \cdot V^{-1} \cdot s
	= V \cdot J^t \cdot x
	\overset{\text{(b)}}{=} V \cdot J^t \cdot B \cdot y
	\overset{\text{(a)}}{=} V \cdot B \cdot J^t \cdot y
	= A \cdot J^t \cdot y
\]

The generalization to the real Jordan decomposition is straightforward by
applying the fact that for complex conjugate eigenvalue pairs $\lambda$ and
$\overline{\lambda}$ the matrix $M$ in a real Jordan block (cf. \cref{omi}) is
similar to the diagonal matrix $D = \left[ \begin{array}{cc}
	\lambda & 0\\
	0 & \overline{\lambda}
\end{array} \right]$ via $U = \left[ \begin{array}{cc}
	-\mathfrak{i} & -\mathfrak{i}\\
	1 & -1
\end{array} \right]$ \citep[Section~3.4.1]{HJ13}, i.e., $M = U \cdot D \cdot
U^{-1}$. The above-mentioned commutativity property~(a) analogously holds for real
Jordan blocks. This completes the proof.

\section{Proof of \cref{exponential}}\label{proof:exponential}

Let $f_k(t)$ denote the value of the $k$-th dimension of $f(t)$, $\lambda$ be
the eigenvalue of $W$ with maximal absolute value and $m$ be the maximal (geometric)
multiplicity of the eigenvalues of the transition matrix $W$. Then, from
\cref{jordan}, we can easily deduce \[ |f_k(t)| = O(t^m\,|\lambda|^t) \]
as asymptotic behavior for large $t$.

\section{Proof of \cref{approx}}\label{proof:approx}

First, we take the series of function values $f(t_0),\dots,f(t_n)$ and identify
them with the time series $S(0),\dots,S(n)$. After applying the LRNN learning
procedure, the LRNN runs through all given values, because by construction the
upper part $\big[ S(0) \cdots S(n-1) \big]$ of the matrix $X$ (cf. \cref{Xin})
and the matrix $Y^\mathrm{out}$ (cf. \cref{Yout}) consist of the series of
function values of $f$, provided that the linear
matrix equation $Y^\mathrm{out} = W^\mathrm{out} \cdot X$ (\cref{linear}) has at
least one solution.

Since \cref{linear} is equivalent to simultaneously solving the equations $y_k =
w_k \cdot X$, for $1 \le k \le N$ where $y_1,\dots,y_N$ and $w_1,\dots,w_N$
denote the row vectors of the matrices $Y^\mathrm{out}$ and $W^\mathrm{out}$,
respectively, the latter is the case if the rank of the coefficient matrix $X$
is equal to the rank of the augmented matrix $M_k = \big[ X ~~ y_k \big]^\top$
for every $k$. This leads to the equation $\mathrm{rank}(X) =
\min(N^\mathrm{in\,out}+N^\mathrm{res},n) = \mathrm{rank}(M_k) =
\min(N^\mathrm{in\,out}+N^\mathrm{res}+1,n)$. From this, it follows that, as
desired, $N^\mathrm{res} \ge n-N^\mathrm{in\,out}$ reservoir neurons have to be
employed to guarantee at least one solution for the $w_k$, provided that the
rank of the matrix $X$ is maximal. For the latter, we consider two cases:
\begin{itemize}
  \item If the rank of the upper part of the matrix $X$ (see above) is not
	maximal, then this does not cause any problems. We only have to replace
	$N^\mathrm{in\,out}$ by the actual rank of the upper part of the matrix
	$X$ in \cref{ineq}.
  \item The rank of the lower part $\big[ R(0) \cdots R(n-1) \big]$ of the
	matrix $X$ almost always has maximal rank, because we employ a random
	reservoir (cf. \cref{thedef}). Thus, a suitable reservoir does the job,
	which completes the proof.
\end{itemize}}

\opt{proof}{\section{Solving Number Puzzles with LLMs}\label{chatgpt}
For comparison, we gave ChatGPT (o3) the task to solve the number puzzles from \citep{GW13}.
For each numbered series below the first seven terms were supplied. The value in
position 8 (in parenthesis) is the ground truth next value. The tasks were (a)
to predict the $8^\text{th}$ value and (b) to give a formula that generates the
entire sequence. The LLM response is shown below the series, split into next
value prediction (which was correct for all 20 sequences), and a suggested
generating formula (usually with alternative solutions). The predicted next
values were correct for all 20 sequences, but some of the suggested generating
formulas were not. Note that the LLM started indexing the values at 0.\\

\noindent
We used the following prompt for the first sequence:
\begin{quote}
\ttfamily
I am interested in time series prediction, and would like to see if it is possible to uncover an underlying function. I will show you the first few values, and you should (a) predict the next value, and (b) tell me the function generating the series.\\

\noindent Let's start with the first series:\\
15,12,8,11,4,7,0
\end{quote}

~\\
We requested \LaTeX{} output for the subsequent series (``\texttt{output the formula in latex code}'') that we used verbatim below. We have annotated mistakes in the formulas in \textcolor{red}{red}, and added remarks in \textcolor{orange}{orange}.

%------------------------------------------------------------
\subsection*{Series \#1 -- \seq{15,\,12,\,8,\,11,\,4,\,7,\,0,\,(3)}}
\begin{enumerate}[label=(\alph*)]
  \item Next value: \nextval{3}.
  \item Rule (recurrence):~ (\textcolor{red}{incorrect formula})
  \[
    a_0=15,\;a_1=12,\quad
    a_n=\begin{cases}
      a_{n-1}+3 & n\text{ odd},\\[4pt]
      a_{n-1}-7 & n\text{ even},\;n\ge2.
    \end{cases}
  \]
\end{enumerate}

%------------------------------------------------------------
\subsection*{Series \#2 -- \seq{148,\,84,\,52,\,36,\,28,\,24,\,22,\,(21)}}
\begin{enumerate}[label=(\alph*)]
  \item Next value: \nextval{21}.
  \item Closed form: 
  \[a_n = 20 + 2^{7-n},\qquad n\ge0.\]
  Recurrence: $a_0=148$ and $a_n=a_{n-1}-2^{7-n}$ for $n\ge1$.
\end{enumerate}

%------------------------------------------------------------
\subsection*{Series \#3 -- \seq{2,\,12,\,21,\,29,\,36,\,42,\,47,\,(51)}}
\begin{enumerate}[label=(\alph*)]
  \item Next value: \nextval{51}.
  \item Closed form:
  \[
    a_n = 2 + 11n - \frac{n(n+1)}{2}=2+\frac{n(21-n)}{2},\qquad n\ge0.
  \]
  Recurrence: $a_0=2$ and $a_{n+1}=a_n+(10-n)$.\\[1em]
  \textcolor{orange}{The alternative formulas (closed form vs recurrence) diverge for larger $n$ but produce the same results for the first 8 relevant values of $n$.} 
\end{enumerate}

%------------------------------------------------------------
\subsection*{Series \#4 -- \seq{2,\,3,\,5,\,9,\,17,\,33,\,65,\,(129)}}
\begin{enumerate}[label=(\alph*)]
  \item Next value: \nextval{129}.
  \item Closed form: $a_n=1+2^{n}$, $n\ge0$.  (Note $a_0=2$.)
  
  Recurrence: $a_{n+1}=2a_n-1$.
\end{enumerate}

%------------------------------------------------------------
\subsection*{Series \#5 -- \seq{2,\,5,\,8,\,11,\,14,\,17,\,20,\,(23)}}
\begin{enumerate}[label=(\alph*)]
  \item Next value: \nextval{23}.
  \item Closed form: $a_n = 2+3n$.
  
  Recurrence: $a_{n+1}=a_n+3$.
\end{enumerate}

%------------------------------------------------------------
\subsection*{Series \#6 -- \seq{2,\,5,\,9,\,19,\,37,\,75,\,149,\,(299)}}
\begin{enumerate}[label=(\alph*)]
  \item Next value: \nextval{299}.
  \item Closed form (\textcolor{red}{incorrect})
  \[
    a_n=\frac{7\,2^{n}-(-1)^{n}}{3},\qquad n\ge0.
  \]
  Recurrence: (\textcolor{red}{incorrect}) \\
  $a_0=2$ and $a_{n+1}=2a_n+(-1)^{n+1}$.
\end{enumerate}

%------------------------------------------------------------
\subsection*{Series \#7 -- \seq{25,\,22,\,19,\,16,\,13,\,10,\,7,\,(4)}}
\begin{enumerate}[label=(\alph*)]
  \item Next value: \nextval{4}.
  \item Formula: $a_n = 25-3n$ \;(equivalently $a_{n+1}=a_n-3$).
\end{enumerate}

%------------------------------------------------------------
\subsection*{Series \#8 -- \seq{28,\,33,\,31,\,36,\,34,\,39,\,37,\,(42)}}
\begin{enumerate}[label=(\alph*)]
  \item Next value: \nextval{42}.
  \item Even--odd definition:
  \[
    a_{2m}=28+3m,\qquad a_{2m+1}=33+3m,\qquad m\ge0.
  \]
  Single‑line version:
  \[a_n = 28+3\lfloor n/2\rfloor+5\,(n\bmod 2).\]
\end{enumerate}

%------------------------------------------------------------
\subsection*{Series \#9 -- \seq{3,\,6,\,12,\,24,\,48,\,96,\,192,\,(384)}}
\begin{enumerate}[label=(\alph*)]
  \item Next value: \nextval{384}.
  \item Closed form: $a_n=3\,2^{n}$. \textcolor{orange}{ correct when interpreted as $3\cdot2^n$}
  
  Recurrence: $a_{n+1}=2a_n$.
\end{enumerate}

%------------------------------------------------------------
\subsection*{Series \#10 -- \seq{3,\,7,\,15,\,31,\,63,\,127,\,255,\,(511)}}
\begin{enumerate}[label=(\alph*)]
  \item Next value: \nextval{511}.
  \item Closed form: $a_n = 2^{n+2}-1$.
  
  Recurrence: $a_{n+1}=2a_n+1$.
\end{enumerate}

%------------------------------------------------------------
\subsection*{Series \#11 -- \seq{4,\,11,\,15,\,26,\,41,\,67,\,108,\,(175)}}
\begin{enumerate}[label=(\alph*)]
  \item Next value: \nextval{175}.
  \item Recurrence:
  \[
    a_0=4,\;a_1=11,\qquad a_{n+1}=a_n+a_{n-1}\quad(n\ge1).
  \]
  Closed form using Fibonacci numbers $F_n$ ($F_0=0,\,F_1=1$):
  \[a_n = 7F_n+4F_{n+1}.\]
\end{enumerate}

%------------------------------------------------------------
\subsection*{Series \#12 -- \seq{5,\,6,\,7,\,8,\,10,\,11,\,14,\,(15)}}
\begin{enumerate}[label=(\alph*)]
  \item Next value: \nextval{15}.
  \item Piece‑wise form:
  \[
    a_{2m}=5+\frac{m(m+3)}{2},\qquad a_{2m+1}=6+\frac{m(m+3)}{2}.
  \]
  Single‑line:
  \[a_n = 5+\frac{\lfloor n/2\rfloor(\lfloor n/2\rfloor+3)}{2}+(n\bmod 2).\]
\end{enumerate}

%------------------------------------------------------------
\subsection*{Series \#13 -- \seq{54,\,48,\,42,\,36,\,30,\,24,\,18,\,(12)}}
\begin{enumerate}[label=(\alph*)]
  \item Next value: \nextval{12}.
  \item Formula: $a_n = 54-6n$ \;(so $a_{n+1}=a_n-6$).
\end{enumerate}

%------------------------------------------------------------
\subsection*{Series \#14 -- \seq{6,\,8,\,5,\,7,\,4,\,6,\,3,\,(5)}}
\begin{enumerate}[label=(\alph*)]
  \item Next value: \nextval{5}.
  \item Recurrence:
  \[
    a_0=6,\qquad a_{n+1}=\begin{cases}a_n+2 & n\text{ even},\\[4pt]a_n-3 & n\text{ odd} .\end{cases}
  \]
  Closed form:
  \[
    a_{2m}=6-m,\qquad a_{2m+1}=8-m.
  \]
  Single‑line:
  \[a_n = 6-\lfloor n/2\rfloor+2\,(n\bmod 2).\]
\end{enumerate}

%------------------------------------------------------------
\subsection*{Series \#15 -- \seq{6,\,9,\,18,\,21,\,42,\,45,\,90,\,(93)}}
\begin{enumerate}[label=(\alph*)]
  \item Next value: \nextval{93}.
  \item Recurrence:
  \[
    a_0=6,\qquad a_{n+1}=\begin{cases}a_n+3 & n\text{ even},\\[4pt]2a_n & n\text{ odd}.\end{cases}
  \]
  Closed form:
  \[
    a_{2m}=12\,2^{m}-6,\qquad a_{2m+1}=12\,2^{m}-3.
  \]
  Single‑line:
  \[a_n = 12\,2^{\lfloor n/2\rfloor}-6+3\,(n\bmod 2).\]
\end{enumerate}

%------------------------------------------------------------
\subsection*{Series \#16 -- \seq{7,\,10,\,9,\,12,\,11,\,14,\,13,\,(16)}}
\begin{enumerate}[label=(\alph*)]
  \item Next value: \nextval{16}.
  \item Recurrence:
  \[
    a_0=7,\quad a_{n+1}=\begin{cases}a_n+3 & n\text{ even},\\[4pt]a_n-1 & n\text{ odd}.\end{cases}
  \]
  Closed form:
  \[
    a_{2m}=7+2m,\qquad a_{2m+1}=10+2m.
  \]
  Single‑line:
  \[a_n = 7+2\lfloor n/2\rfloor+3\,(n\bmod 2).\]
\end{enumerate}

%------------------------------------------------------------
\subsection*{Series \#17 -- \seq{8,\,10,\,14,\,18,\,26,\,34,\,50,\,(66)}}
\begin{enumerate}[label=(\alph*)]
  \item Next value: \nextval{66}.
  \item Recurrence:
  \[
    a_0=8,\qquad a_{n+1}=a_n+2^{\left\lfloor\frac{n+3}{2}\right\rfloor}.
  \]
  Closed form (even/odd):
  \[
    a_{2m}=3\,2^{m+1}+2,\qquad a_{2m+1}=4\,2^{m+1}+2.
  \]
  Single‑line:
  \[a_n = 2\Bigl((3+(n\bmod 2))\,2^{\lfloor n/2\rfloor}+1\Bigr).\]
\end{enumerate}

%------------------------------------------------------------
\subsection*{Series \#18 -- \seq{8,\,12,\,10,\,16,\,12,\,20,\,14,\,(24)}}
\begin{enumerate}[label=(\alph*)]
  \item Next value: \nextval{24}.
  \item Closed form:
  \[
    a_{2k}=8+2k,\qquad a_{2k+1}=12+4k.
  \]
  Single‑line:
  \[a_n = 8+2\lfloor n/2\rfloor+(n\bmod 2)\bigl(4+2\lfloor n/2\rfloor\bigr).\]
\end{enumerate}

%------------------------------------------------------------
\subsection*{Series \#19 -- \seq{8,\,12,\,16,\,20,\,24,\,28,\,32,\,(36)}}
\begin{enumerate}[label=(\alph*)]
  \item Next value: \nextval{36}.
  \item Formula: $a_n=8+4n$ \;(so $a_{n+1}=a_n+4$).
\end{enumerate}

%------------------------------------------------------------
\subsection*{Series \#20 -- \seq{9,\,20,\,6,\,17,\,3,\,14,\,0,\,(11)}}
\begin{enumerate}[label=(\alph*)]
  \item Next value: \nextval{11}.
  \item Even--odd closed form:
  \[
    a_{2k}=9-3k,\qquad a_{2k+1}=20-3k,\qquad k\ge0.
  \]
  Single‑line variant:
  \[a_n = 9+11\left\lceil\tfrac{n}{2}\right\rceil-14\left\lfloor\tfrac{n}{2}\right\rfloor.\]
  Recurrence:
  \[
    a_0=9,\qquad a_{n+1}=\begin{cases}a_n+11 & n\text{ even},\\[4pt]a_n-14 & n\text{ odd}.\end{cases}
  \]
\end{enumerate}}

\end{document}